\documentclass[3p,times]{article}

\usepackage{arxiv}

\usepackage{amssymb}
\usepackage{amsthm}
\usepackage{mathtools}

\usepackage{bm} 
\usepackage{upgreek} 
\usepackage{amsmath}
\usepackage{makecell}

\usepackage{url}

\usepackage{subcaption}
\usepackage{enumitem}
\usepackage{algpseudocodex}
\usepackage{algorithm}
\algrenewcommand\algorithmicrequire{\textbf{Input:}}
\algrenewcommand\algorithmicensure{\textbf{Output:}}

\usepackage{booktabs}       

\theoremstyle{definition}
\newtheorem{definition}{Definition}

\newtheorem{proposition}{Proposition}

\theoremstyle{plain}
\newtheorem{theorem}{Theorem}
\newtheorem{corollary}{Corollary}
\newtheorem{lemma}{Lemma}

\usepackage{mathtools}
\newcommand{\inp}[2]{{#1^\top}{#2}}
\DeclarePairedDelimiterX{\dotinv}[2]{\big(}{\big)}{#1^\top#2}
\DeclarePairedDelimiterX{\augm}[2]{\big(}{\big)}{#1 \big| #2}
\DeclarePairedDelimiter{\norm}{\lVert}{\rVert}

\DeclareMathOperator*{\argmax}{arg\,max}
\DeclareMathOperator{\tr}{tr}
\DeclareMathOperator{\svd}{svd}
\DeclareMathOperator{\orth}{orth}
\DeclareMathOperator{\normalise}{normalise}
\newenvironment{amatrix}[1]{%
  \left(\begin{array}{@{}*{#1}{c}|c@{}}
}{%
  \end{array}\right)
}

\newcommand \RV[1]{{#1}}
\newcommand \HL[1]{{#1}}
\usepackage{authblk}

\title{Canonical-Correlation-Based Fast Feature Selection for Structural Health Monitoring}

\author[2]{Sikai Zhang}
\author[1,3]{Tingna Wang\thanks{Corresponding author. \texttt{tina\_wang@tongji.edu.cn}}}
\author[2]{Keith Worden}
\author[1,3,4]{Limin Sun}
\author[2]{Elizabeth J. Cross}

\affil[1]{Department of Bridge Engineering, Tongji University, Shanghai, China}
\affil[2]{Dynamics Research Group, Department of Mechanical Engineering, The University of Sheffield, Sheffield, UK}
\affil[3]{Shanghai Qizhi Institute, Shanghai, China}
\affil[4]{State Key Laboratory of Disaster Reduction in Civil Engineering, Tongji University, Shanghai, China}
\begin{document}
\maketitle

\begin{abstract}
Feature selection refers to the process of selecting useful features for machine learning tasks, and it is also a key step for structural health monitoring (SHM).
This paper proposes a fast feature selection algorithm by efficiently computing the sum of squared canonical correlation coefficients between monitored features and target variables of interest in greedy search.
The proposed algorithm is applied to both synthetic and real datasets to illustrate its advantages in terms of computational speed, general classification and regression tasks, as well as damage-sensitive feature selection tasks.
\RV{Furthermore, the performance of the proposed algorithm is evaluated under varying environmental conditions and on an edge computing device to investigate its applicability in real-world SHM scenarios.}
The results show that the proposed algorithm can successfully select useful features with extraordinarily fast computational speed, which implies that the proposed algorithm has great potential where features need to be selected and updated online frequently, or where devices have limited computing capability.
\end{abstract}


\section{Introduction}
Feature selection is an important step in machine learning to determine a useful subset of features for the construction of a model \cite{guyon2003introduction}.
It is widely used in the fields of structural health monitoring (SHM) \RV{based on the statistical pattern recognition}, such as damage-sensitive feature selection \cite{rutherford2007non,zhang2022vibration,hughes2021probabilistic}, sensor placement optimisation \cite{barthorpe2020emerging,wang2022assessment,yang2022optimal,wang2022improved}, query selection for active learning \cite{bull2019probabilistic,hughes2022risk,bull2020towards,hughes2022robust}, and equation discovery \cite{fuentes2021equation}. 
\RV{In addition to these researches that use feature selection as a step in building an SHM system, some efforts have been made to specifically utilise feature selection to solve some challenging problems in SHM. 
A signal simulation-based feature selection algorithm was established in \cite{gharehbaghi2021deterioration} to find the most sensitive and uncorrelated features by only using the healthy state data of a structure. This algorithm can handle the common situation in the SHM field where there is no damage-state data \cite{sarmadi2023probabilistic}.
In \cite{alves2023automated}, the unsupervised infinite feature selection method was used to filter out the effects of changing traffic and environmental variations to some extent in damage localisation tasks. 
To cope with the data scarcity problem, a LASSO-based wrapper method was developed in \cite{bee2024multitask} to facilitate feature knowledge transfer between structures. This wrapper method can provide more general and meaningful features by using patterns shared across different tasks. 
This paper does not aim to provide feature selection techniques tailored to address specific challenges in the SHM domain, but rather to provide a comprehensive and efficient method to select useful features for machine learning-based SHM from a more general perspective.}

Feature selection methods can generally be categorised into filter, wrapper, and embedded methods \cite{guyon2003introduction}.
Filter methods select features by model-training-free ranking criteria (e.g., Mutual Information (MI) and correlation coefficients) to evaluate the effectiveness of features.
For wrapper methods, the effectiveness of features is evaluated by the predictive performance of machine learning models.
Embedded methods select features during parameter optimisation by manipulating the objective function, such as Lasso \cite{tibshirani1996regression} or the structure of a model, such as CART \cite{breiman1984classification}.

Unlike wrapper and embedded methods, filter methods do not involve model training, which makes them faster than the other two methods.
However, filter methods are also different from the other two methods in that they usually evaluate a single feature at a time without considering redundancy and interactions between features \cite{li2017recent}.
The problem with ignoring redundancy is that some selected features are highly ranked but provide similar information.
The problem with not considering interaction is that some features that work well with others are ignored because they are not favourable on their own, e.g., the inputs of an XOR gate.

The mainstream strategy to overcome the aforementioned two issues is to propose a new ranking criterion by quantifying the feature relevance, the feature redundancy, and the feature interaction, respectively.
The correlation-based feature selection method \cite{hall1999feature} uses the averaged feature-feature Pearson's correlation coefficients as the redundancy to penalise the averaged feature-class Pearson's correlation coefficients.
The minimal-Redundancy-Maximal-Relevance (mRMR) method \cite{peng2005feature} uses averaged feature-feature MI as the redundancy to penalise the averaged feature-class MI.
The unifying framework proposed in \cite{brown2012conditional} uses MI-based criteria to define feature relevancy, redundancy and conditional redundancy (i.e. interaction), and the feature relevancy is penalised by feature redundancy and compensated by feature conditional redundancy.
Another strategy that receives much less attention is to evaluate the overall relevance between multiple features and the target with a variable, such as coefficient of determination, canonical correlation coefficient \cite{kaya2014cca} and conditional MI \cite{brown2012conditional}.
\HL{Since these criteria evaluate a feature subset directly, they are inherently immune to information redundancy and interaction issues \RV{to some extent}.
At this point, it can be realised that an attractive research direction is to develop a filter feature selection method, which can evaluate a set of features at a time like wrapper and embedded methods, but as fast as traditional filter methods.}

\RV{Based on the discussion above, this} paper adopts the second strategy, using the Sum of Squared canonical Correlation coefficients (SSC) as the ranking criterion to evaluate feature subsets \RV{via a filter way}. 
The effectiveness of the SSC has been empirically demonstrated in classification cases in \cite{zhang2022orthogonal}\RV{, but this criterion is not very attractive in terms of computing speed.} \RV{Therefore, an orthogonalisation method combined with greedy search was adopted in \cite{zhang2022orthogonal} to speed up the computational speed of the SSC-based filter method.}
The idea of accelerating the calculation of a ranking criterion in greedy search by orthogonalising the feature matrix was first introduced in \cite{korenberg1989robust} to select terms for Nonlinear AutoRegressive with eXogenous input (NARX) models. 
Another research \cite{chen1989orthogonal} then adopted the same idea to select terms for Nonlinear AutoRegressive Moving Average with eXogenous input (NARMAX) models by computing Error Reduction Ratios (ERR).
In \cite{stoppiglia2003ranking}, the ERR-based method was introduced to the machine learning area for feature selection.
However, all these methods can only accept one variable as the target.
To cover the case where the number of target variables exceeds one, \RV{the {$h$-correlation}-based method was proposed in \cite{zhang2022orthogonal} to increase the calculation speed of the SSC criterion in greedy search by orthogonalising the feature matrix.} 
\RV{Inspired by the work in \cite{zhang2022orthogonal}, this paper proposes the \textit{$\eta$-cosine}-based method to achieve the same functionality as the $h$-correlation-based method, but it is faster than the $h$-correlation-based method when the number of instances exceeds the number of features.
Note that the features selected by the two methods should be the same. The difference between these two methods lies in the calculation process of SSC.}

\RV{The major novelties of this paper include: 
\begin{enumerate}
    \item This paper proposes another fast method to select features using SSC as a ranking criterion. A new statistic \textit{$\eta$-cosine} is defined to better describe intermediate quantities in the calculation process.
    \item The filter method based on \textit{$\eta$-cosine} and the filter method based on $h$-correlation are unified into a feature selection algorithm based on SSC to select useful features for different situations efficiently. The corresponding theorems supporting this unification are also provided.
    \item The speed and performance advantages of the proposed algorithm are validated on open-access datasets and two real structures, ensuring its effectiveness and applicability in real-world scenarios.    
    \item The proposed algorithm is deployed on an edge computing module and tested.
\end{enumerate}
}

Section \ref{s:corr_angle} briefly introduces three correlation coefficients and their corresponding angles as the prerequisite knowledge.
The supporting theorems for \RV{$h$-correlation- and $\eta$-cosine-based methods} are developed in Section \ref{s:theorem}.
Although \RV{the $h$-correlation-based filter method} is not new, its supporting theorems here are more concise and rigorous, which also helps to clearly show the relationship with the proposed \RV{$\eta$-cosine-based method}.
In Section \ref{s:greedy}, \RV{$h$-correlation- and $\eta$-cosine-based methods} are unified into a fast SSC-based feature selection algorithm.
\HL{\RV{For the sake of brevity,} hereafter the fast SSC-based feature selection algorithm using $h$-correlation for acceleration is referred to as the \textit{$h$-correlation method}, while using \textit{$\eta$-cosine} for acceleration is referred to as the \textit{$\eta$-cosine method}.}
In Section \ref{s:case studies}, a synthetic dataset is used to demonstrate the speed advantage of the $h$-correlation method and the $\eta$-cosine method.
Then, eight real datasets are applied to \RV{compare filter methods with different feature-ranking criteria in} classification and regression tasks.
\RV{Finally, three case studies are conducted using datasets from two aircraft and a simulated dataset (on an edge computing module) to provide practical examples for the proposed algorithm in SHM applications.}
The conclusions of this paper are at the end.

\section{Correlation coefficients and angles}\label{s:corr_angle}
This section summarises the basic and necessary knowledge about the definitions of correlation coefficients and angles to increase the independence and readability of this paper.

The \textit{Pearson's correlation coefficient} is a criterion to measure the linear association between two random variables.
If the vector $\mathbf{x} \in \mathbb{R}^{N\times 1}$ contains the samples of one variable and the vector $\mathbf{y} \in \mathbb{R}^{N\times 1}$ contains the samples of another variable, the association between $\mathbf{x}$ and $\mathbf{y}$ can be measured by the Pearson's correlation coefficient $r({\mathbf{x},\mathbf{y}})$ which is given by,
\begin{equation*}
    r({\mathbf{x},\mathbf{y}}) \coloneqq \frac{\inp{\mathbf{x}_\mathrm{C}}{\mathbf{y}_\mathrm{C}}}{\norm{\mathbf{x}_\mathrm{C}}\;\norm{\mathbf{y}_\mathrm{C}}}
,\end{equation*}
where the operator $\coloneqq$ denotes a definition, the operator $\norm{\cdot}$ denotes a $\ell^2$ norm, $\mathbf{x}_\mathrm{C}$ is the centred $\mathbf{x}$ given by,
\begin{equation*}
    \mathbf{x}_\mathrm{C} = \begin{pmatrix}
    x_1 - \bar{x} \\
    \vdots\\
    x_n - \bar{x} \\
    \end{pmatrix},
    \quad \bar{x} = \sum_{i=1}^{N} x_i /N,
    \quad x_i \in \mathbb{R}
,\end{equation*}
and $\mathbf{y}_\mathrm{C}$ are the centred $\mathbf{y}$.
A criterion similar to Pearson's correlation coefficient is the cosine of the angle between the column vectors $\mathbf{x}$ and $\mathbf{y}$, i.e. $\cos(\angle(\mathbf{x}, \mathbf{y}))$, where,
\begin{equation*} 
     \angle(\mathbf{x}, \mathbf{y}) \coloneqq \arccos\left(\frac{\inp{\mathbf{x}} {\mathbf{y}}}{\norm{\mathbf{x}}\; \norm{\mathbf{y}}}\right) \in [0,\pi]
.\end{equation*}
Unlike Pearson's correlation coefficient, the calculation of $\cos(\angle(\mathbf{x}, \mathbf{y}))$ does not require one to centre the vectors.

The \textit{multiple correlation coefficient} is a criterion to measure the linear association between a multivariate random variable and a random variable \cite{cohen2014applied}.
If the design matrix $\mathbf{X} \in \mathbb{R}^{N\times n}$ contains the samples of $n$ variables and the target vector $\mathbf{y} \in \mathbb{R}^{N\times 1}$ contains the samples of a variable, the association between $\mathbf{X}$ and $\mathbf{y}$ can be measured by the multiple correlation coefficient $R({\mathbf{X},\mathbf{y}})$ or $R({\mathbf{y},\mathbf{X}})$.
The multiple correlation coefficient between $\mathbf{X}$ and $\mathbf{y}$ is determined by finding a projection direction for $\mathbf{X}_\mathrm{C}$, so that the Pearson's correlation coefficient between $\mathbf{y}_\mathrm{C}$ and the projected $\mathbf{X}_\mathrm{C}$ is maximised, i.e.
\begin{equation*} 
    R({\mathbf{X},\mathbf{y}}) = R({\mathbf{y},\mathbf{X}}) \coloneqq \max_{\bm{\upalpha}} r({\mathbf{X}_\mathrm{C}\bm{\upalpha},\mathbf{y}_\mathrm{C}})
,\end{equation*}
where $\mathbf{X}_\mathrm{C}$ is the column-centred $\mathbf{X}$ given by,
\begin{equation*}
    \mathbf{X}_\mathrm{C} = \begin{pmatrix}
    x_{1,1} - \bar{x}_1 & \hdots & x_{1,n} - \bar{x}_n \\
    \vdots & \ddots & \vdots \\
    x_{N,1} - \bar{x}_1 & \hdots & x_{N,n} - \bar{x}_n
    \end{pmatrix},
    \quad \bar{x}_j = \sum_{i=1}^{N} x_{i,j} /N,
    \quad x_{i,j} \in \mathbb{R}
,\end{equation*}
$\mathbf{y}_\mathrm{C}$ is the centred $\mathbf{y}$, and the optimal projection direction $\bm{\upalpha} \in \mathbb{R}^n$ is given by,
\begin{equation*} 
    \bm{\upalpha} = \dotinv{\mathbf{X}_\mathrm{C}}{\mathbf{X}_\mathrm{C}}^{-1}\inp{\mathbf{X}_\mathrm{C}}{\mathbf{y}_\mathrm{C}}
.\end{equation*}
Correspondingly, a criterion similar to multiple correlation coefficient is the maximised cosine of the angle between the projected $\mathbf{X}$ and $\mathbf{y}$
\begin{equation*}
    \max_{\bm{\upalpha}}\cos{\big(\angle(\mathbf{X}\bm{\upalpha}, \mathbf{y})\big)} =\cos{\big(\min_{\bm{\upalpha}} \angle(\mathbf{X}\bm{\upalpha}, \mathbf{y})\big)}
.\end{equation*}
Accordingly, the angle between $\mathbf{X}$ and $\mathbf{y}$ can be defined as,
\begin{equation} \label{eq:mv_angle}
     \varTheta(\mathbf{X}, \mathbf{y}) = \varTheta(\mathbf{y}, \mathbf{X}) \coloneqq \min_{\bm{\upalpha}} \angle(\mathbf{X}\bm{\upalpha}, \mathbf{y})
,\end{equation}
where the optimal projection $\bm{\upalpha} \in \mathbb{R}^n$ is given by,
\begin{equation*} 
    \bm{\upalpha} = \dotinv{\mathbf{X}}{\mathbf{X}}^{-1}\inp{\mathbf{X}}{\mathbf{y}}
.\end{equation*}

The \textit{canonical correlation coefficient} is a criterion to measure the linear association between two multivariate random variables \cite{harold1936relations,hardoon2004canonical}.
If the design matrix $\mathbf{X} \in \mathbb{R}^{N\times n}$ contains the samples of $n$ variables, the target matrix $\mathbf{Y} \in \mathbb{R}^{N\times m}$ contains the samples of $m$ variables, the association between $\mathbf{X}$ and $\mathbf{Y}$ can be measured by the canonical correlation coefficient $R_i({\mathbf{X},\mathbf{Y}})$.
\textit{Canonical Correlation Analysis} is a way of computing $R_i({\mathbf{X},\mathbf{Y}})$ by finding pairs of the projection directions $\bm{\upalpha}_i$ and $\bm{\upbeta}_i$ to make the Pearson's correlation coefficient between $\mathbf{X}_\mathrm{C}{\bm{\upalpha}_i}$ and $\mathbf{Y}_\mathrm{C}{\bm{\upbeta}_i}$ for the $i\textsuperscript{th}$ pair of projection direction is maximised with all projections orthogonal to each other, i.e.
\begin{subequations} \label{eq:defcc}
\begin{align}
    R_i({\mathbf{X},\mathbf{Y}}) &\coloneqq \max_{\bm{\upalpha}_i,\bm{\upbeta}_i} r({\mathbf{X}_\mathrm{C}\bm{\upalpha}_i,\mathbf{Y}_\mathrm{C}\bm{\upbeta}_i})\\
    \shortintertext{subject to}
    \inp{(\mathbf{X}_\mathrm{C}\bm{\upalpha}_i)}{(\mathbf{X}_\mathrm{C}\bm{\upalpha}_j)}&=0 \quad \text{for} \quad i\neq j,\\
    \inp{(\mathbf{Y}_\mathrm{C}\bm{\upbeta}_i)}{(\mathbf{Y}_\mathrm{C}\bm{\upbeta}_j)}&=0 \quad \text{for} \quad i\neq j
,\end{align}
\end{subequations}
where $\mathbf{X}_\mathrm{C}$ and $\mathbf{Y}_\mathrm{C}$ are the column-centred $\mathbf{X}$ and $\mathbf{Y}$. 
The optimal vectors $\bm{\upalpha}_i \in \mathbb{R}^{n\times 1}$ and $\bm{\upbeta}_i \in \mathbb{R}^{m\times 1}$ are obtained by solving the eigenvalue problem given by \cite{hardoon2004canonical},
\begin{subequations}\label{eq:ccc0}
\begin{align}
    \dotinv{\mathbf{X}_\mathrm{C}}{\mathbf{X}_\mathrm{C}}^{-1} \inp{\mathbf{X}_\mathrm{C}}{\mathbf{Y}_\mathrm{C}} \dotinv{\mathbf{Y}_\mathrm{C}}{\mathbf{Y}_\mathrm{C}}^{-1} \inp{\mathbf{Y}_\mathrm{C}}{\mathbf{X}_\mathrm{C}} \bm{\upalpha}_i &= R_i^2({\mathbf{X},\mathbf{Y}}) \bm{\upalpha}_i \label{eq:ccca}\\
    \dotinv{\mathbf{Y}_\mathrm{C}}{\mathbf{Y}_\mathrm{C}}^{-1} \inp{\mathbf{Y}_\mathrm{C}}{\mathbf{X}_\mathrm{C}} \dotinv{\mathbf{X}_\mathrm{C}}{\mathbf{X}_\mathrm{C}}^{-1} \inp{\mathbf{X}_\mathrm{C}}{\mathbf{Y}_\mathrm{C}} \bm{\upbeta}_i &= R_i^2({\mathbf{X},\mathbf{Y}}) \bm{\upbeta}_i \label{eq:cccb}
.\end{align}
\end{subequations}
The two projection directions $\bm{\upalpha}_i$ and $\bm{\upbeta}_i$ are the eigenvectors, and the corresponding eigenvalue is the squared canonical correlation coefficient $R_i^2({\mathbf{X},\mathbf{Y}})$.
If $\mathbf{X}_\mathrm{C}$ and $\mathbf{Y}_\mathrm{C}$ have full column rank, the number of non-zero eigenvalues of \eqref{eq:ccc0} is not more than $n \wedge m$, where the operator $\wedge$ returns the minimum of the two arguments.
Thus, in contrast with the multiple correlation coefficient\textemdash which is a single value\textemdash the canonical correlation coefficients for $\mathbf{X}$ and $\mathbf{Y}$ include $n \wedge m$ values, which are denoted as $R_1({\mathbf{X},\mathbf{Y}}),\ldots,R_{n \wedge m}({\mathbf{X},\mathbf{Y}})$.
The multiple correlation is a special case of the canonical correlation when $\mathbf{X}$ or $\mathbf{Y}$ is a vector.
The corresponding cosine of the $i\textsuperscript{th}$ angle between the projected $\mathbf{X}$ and the projected $\mathbf{Y}$ is,
\begin{subequations} \label{eq:defcos}
\begin{align}
    \max_{\bm{\upalpha}_i,\bm{\upbeta}_i}\cos{\big(\angle(\mathbf{X}\bm{\upalpha}_i, \mathbf{Y}\bm{\upbeta}_i)\big)} &=\cos{\big(\min_{\bm{\upalpha}_i,\bm{\upbeta}_i} \angle(\mathbf{X}\bm{\upalpha}_i, \mathbf{Y}\bm{\upbeta}_i)\big)}\\
    \shortintertext{subject to}
    \inp{(\mathbf{X}\bm{\upalpha}_i)}{(\mathbf{X}\bm{\upalpha}_j)}&=0 \quad \text{for} \quad i\neq j,\\
    \inp{(\mathbf{Y}\bm{\upbeta}_i)}{(\mathbf{Y}\bm{\upbeta}_j)}&=0 \quad \text{for} \quad i\neq j
.\end{align}
\end{subequations}
Accordingly, the $i\textsuperscript{th}$ angle between $\mathbf{X}$ and $\mathbf{Y}$ can be defined as,
\begin{subequations} \label{eq:mm_angle}
\begin{align}
    \varTheta_i(\mathbf{X}, \mathbf{Y}) &\coloneqq \min_{\bm{\upalpha}_i,\bm{\upbeta}_i}     \angle(\mathbf{X}\bm{\upalpha}_i, \mathbf{Y}\bm{\upbeta}_i)
    \shortintertext{subject to}
    \inp{(\mathbf{X}\bm{\upalpha}_i)}{(\mathbf{X}\bm{\upalpha}_j)}&=0 \quad \text{for} \quad i\neq j,\\
    \inp{(\mathbf{Y}\bm{\upbeta}_i)}{(\mathbf{Y}\bm{\upbeta}_j)}&=0 \quad \text{for} \quad i\neq j
,\end{align}
\end{subequations}
where the optimal projections $\bm{\upalpha}_i \in \mathbb{R}^{n\times 1}$ and $\bm{\upbeta}_i \in \mathbb{R}^{m\times 1}$ are given by,
\begin{subequations}\label{eq:angle_eig}
\begin{align} 
    \dotinv{\mathbf{X}}{\mathbf{X}}^{-1} \inp{\mathbf{X}}{\mathbf{Y}} \dotinv{\mathbf{Y}}{\mathbf{Y}}^{-1} \inp{\mathbf{Y}}{\mathbf{X}} \bm{\upalpha}_i &= \cos^2\big(\varTheta_i({\mathbf{X},\mathbf{Y}})\big) \bm{\upalpha}_i\\
    \dotinv{\mathbf{Y}}{\mathbf{Y}}^{-1} \inp{\mathbf{Y}}{\mathbf{X}} \dotinv{\mathbf{X}}{\mathbf{X}}^{-1} \inp{\mathbf{X}}{\mathbf{Y}} \bm{\upbeta}_i &= \cos^2\big(\varTheta_i({\mathbf{X},\mathbf{Y}})\big) \bm{\upbeta}_i
.\end{align}
\end{subequations}
The $n \wedge m$ minimised angles are denoted as $\varTheta_1({\mathbf{X},\mathbf{Y}}),\ldots,\varTheta_{n \wedge m}({\mathbf{X},\mathbf{Y}})$, which are referred to as the \textit{principal angles} \cite{golub2013matrix}.
The angle defined in \eqref{eq:mv_angle} is a special case of the angle defined in \eqref{eq:mm_angle}, when $\mathbf{X}$ or $\mathbf{Y}$ is a vector.

\section{Theorems of canonical correlation} \label{s:theorem}
The following definition is given for developing the theorems of canonical correlation required for $h$-correlation and $\eta$-cosine.

\begin{definition}\label{d:1}
Let $\mathbf{U} \in \mathbb{R}^{N\times n}$ be a matrix where all columns are linearly independent, called a vector basis.
If the columns of $\mathbf{X} \in \mathbb{R}^{N\times p}$ are in the range of $\mathbf{U}$, then there is one and only one matrix $[\mathbf{X}]_{\mathbf{U}} \in \mathbb{R}^{n\times p}$ satisfying,
\begin{equation*}
    \mathbf{X} = \mathbf{U}[\mathbf{X}]_{\mathbf{U}}
,\end{equation*}
and the matrix $[\mathbf{X}]_{\mathbf{U}}$ is defined as the \textbf{coordinate matrix} of $\mathbf{X}$ with respect to $\mathbf{U}$.
\end{definition}

Based on Definition \ref{d:1}, it is straightforward to obtain the following proposition.
\begin{proposition} \label{p:1}
If $[\mathbf{X}]_{\mathbf{U}}$ and $[\mathbf{Y}]_{\mathbf{U}}$ are the coordinate matrices of $\mathbf{X}$ and $\mathbf{Y}$ with respect to the same matrix $\mathbf{U}$, then,
\begin{equation*}
    \augm{[\mathbf{X}]_{\mathbf{U}}}{[\mathbf{Y}]_{\mathbf{U}}} = \big[\augm{\mathbf{X}}{\mathbf{Y}}\big]_{\mathbf{U}}
,\end{equation*}
where $\augm{\cdot}{\cdot}$ denotes an augmented matrix.
\end{proposition}

By finding some special vector bases, the canonical correlation coefficient will have a superposition property, which is the key to speeding up the SSC-based feature selection algorithm.

\begin{theorem}[Correlation Superposition Theorem] \label{c:1}
If, 
\begin{enumerate}[label=\textbf{\emph{T\ref{c:1}.\arabic*}},leftmargin=*,align=left]
    \item\label{s:c1.b} $\mathbf{X}_\mathrm{C}$ is the column-centred matrix of $\mathbf{X} \in \mathbb{R}^{N\times n}$;
    \item $\mathbf{Y} \in \mathbb{R}^{N\times m}$ is a full column rank matrix;
    \item\label{s:c1.1} the columns of $\mathbf{X}_\mathrm{C}$ are in the range of $\mathbf{W} \in \mathbb{R}^{N\times n}$ whose columns are centred and form a vector basis;
    \item $\mathbf{W} = \augm{\mathbf{W}_\mathrm{s}}{\mathbf{W}_\mathrm{r}}$, where $\mathbf{W}_\mathrm{s} \in \mathbb{R}^{N\times p}$ and $\mathbf{W}_\mathrm{r} \in \mathbb{R}^{N\times q}$ and $n = p+q$;
    \item\label{s:c1.2} $\mathbf{W}_\mathrm{r}$ is orthogonal to $\mathbf{W}_\mathrm{s}$, i.e. $\inp{\mathbf{W}_\mathrm{s}}{\mathbf{W}_\mathrm{r}} = \mathbf{0}_{p\times q}$, where $\mathbf{0}_{p\times q}$ is a p-by-q zero matrix,
\end{enumerate}
then,
\begin{equation} \label{eq:c1}
    \sum_{k = 1}^{n \wedge m}R_k^2\big({\mathbf{X},\mathbf{Y}}\big) = \sum_{k = 1}^{n \wedge m}R_k^2\big({\mathbf{W},\mathbf{Y}}\big) = \sum_{k = 1}^{p \wedge m}R_k^2({\mathbf{W}_\mathrm{s},\mathbf{Y}}) + \sum_{k = 1}^{q \wedge m}R_k^2({\mathbf{W}_\mathrm{r},\mathbf{Y}})
.\end{equation}
\end{theorem}
\emph{\ref{s:c1.b}} to \emph{\ref{s:c1.2}} indicate the assumptions applied in Theorem \ref{c:1}, where the assumption \emph{\ref{s:c1.2}} can be satisfied by the classical Gram-Schmidt process or the modified Gram-Schmidt process \cite{golub2013matrix}.
The proof of Theorem \ref{c:1} is given in \ref{ap:a}.
Using the Correlation Superposition Theorem, the following corollary can be obtained.
\begin{corollary}[Maximum Correlation Theorem] \label{t:1}
If,
\begin{enumerate}[label=\textbf{\emph{C\ref{t:1}.\arabic*}},leftmargin=*,align=left]
    \item\label{s:1.b} $\mathbf{X}_\mathrm{C}$ and $\mathbf{Y}_\mathrm{C}$ are the column-centred matrices of $\mathbf{X} \in \mathbb{R}^{N\times n}$ and $\mathbf{Y} \in \mathbb{R}^{N\times m}$;
    \item\label{s:1.1} the columns of $\mathbf{X}_\mathrm{C}$ are in the range of $\mathbf{W} \in \mathbb{R}^{N\times n}$ whose columns are centred and form an orthogonal basis, where $\mathbf{W} = (\mathbf{w}_{1}|\mathbf{w}_{2}|\ldots|\mathbf{w}_{n})$ and $(\cdot|\cdot|\ldots|\cdot)$ denotes an augmented matrix;
    \item\label{s:1.2} the columns of $\mathbf{Y}_\mathrm{C}$ are in the range of $\mathbf{V} \in \mathbb{R}^{N\times m}$ whose columns are centred and form an orthogonal basis, where $\mathbf{V} = (\mathbf{v}_{1}|\mathbf{v}_{2}|\ldots|\mathbf{v}_{m})$,
\end{enumerate}
then,
\begin{equation} \label{eq:t1}
    \sum_{k = 1}^{n \wedge m}R_k^2({\mathbf{X},\mathbf{Y}}) = \sum_{k = 1}^{n \wedge m}R_k^2({\mathbf{W},\mathbf{V}}) = \sum_{i = 1}^{n}\sum_{j = 1}^{m}h_{i,j}
,\end{equation}
where,
\begin{equation} \label{eq:h}
    h_{i,j} = r^2(\mathbf{w}_{i},\mathbf{v}_{j})
.\end{equation}
\end{corollary}
\emph{\ref{s:1.b}} to \emph{\ref{s:1.2}} indicate the assumptions applied in Corollary \ref{t:1}.
It is seen that the maximisation in \eqref{eq:defcc} can be decomposed to the orthogonalisation and summation.
On this basis, the $h$-correlation method decomposes the SSC to $h$-correlations \eqref{eq:h} to speed up its computation, which will be explained in Section \ref{s:greedy}.

Then, the lemma connecting the $h$-correlation method and the $\eta$-cosine method is given as,
\begin{lemma} \label{lemma:1}
If, 
\begin{enumerate}[label=\textbf{\emph{L\ref{lemma:1}.\arabic*}},leftmargin=*,align=left]
    \item\label{s:l1.b} $\mathbf{X}_\mathrm{C}$ and $\mathbf{Y}_\mathrm{C}$ are the column-centred matrices of $\mathbf{X} \in \mathbb{R}^{N\times n}$ and $\mathbf{Y} \in \mathbb{R}^{N\times m}$;
    \item\label{s:l1.1} $[\mathbf{X}_\mathrm{C}]_{\mathbf{U}}$ and $[\mathbf{Y}_\mathrm{C}]_{\mathbf{U}}$ are the coordinate matrices of $\mathbf{X}_\mathrm{C}$ and $\mathbf{Y}_\mathrm{C}$ with respect to the same matrix $\mathbf{U} \in \mathbb{R}^{N\times (n+m)}$ whose columns form an orthonormal basis, where `orthonormal' means $\inp{\mathbf{U}}{\mathbf{U}} = \mathbf{I}_{n+m}$ and $\mathbf{I}_{n+m}$ is a (n+m)-by-(n+m) identity matrix,
\end{enumerate}
then,
\begin{equation*}
    R_i(\mathbf{X}, \mathbf{Y}) = \cos\big(\varTheta_i([\mathbf{X}_\mathrm{C}]_{\mathbf{U}}, [\mathbf{Y}_\mathrm{C}]_{\mathbf{U}})\big) \quad \text{for} \quad i=1,\ldots,n \wedge m
.\end{equation*}
\end{lemma}
\emph{\ref{s:l1.b}} to \emph{\ref{s:l1.1}} indicate the assumptions applied in Lemma \ref{lemma:1}.
The proof is given in \ref{ap:b}.
Using this lemma, the canonical correlation coefficient between two matrices is transferred to the angle between their coordinate matrices concerning the same orthonormal matrix. When $N \gg n+m$, the computation speed of canonical correlation can be significantly improved via this lemma.
Moreover, the lemma is the key to transitioning from the $h$-correlation method and the $\eta$-cosine method.
The theorems of canonical correlation required for the proposed $\eta$-cosine method are developed as follows. 

\begin{theorem}[Cosine Superposition Theorem] \label{c:2}
If, 
\begin{enumerate}[label=\textbf{\emph{T\ref{c:2}.\arabic*}},leftmargin=*,align=left]
    \item\label{s:c2.b} $\mathbf{X}_\mathrm{C}$ and $\mathbf{Y}_\mathrm{C}$ are the column-centred matrices of $\mathbf{X} \in \mathbb{R}^{N\times n}$ and $\mathbf{Y} \in \mathbb{R}^{N\times m}$;
    \item\label{s:c2.0} $[\mathbf{X}_\mathrm{C}]_{\mathbf{U}}$ and $[\mathbf{Y}_\mathrm{C}]_{\mathbf{U}}$ are the coordinate matrices of $\mathbf{X}_\mathrm{C}$ and $\mathbf{Y}_\mathrm{C}$ with respect to the same matrix $\mathbf{U} \in \mathbb{R}^{N\times (n+m)}$ whose columns form an orthonormal basis, where `orthonormal' means $\inp{\mathbf{U}}{\mathbf{U}} = \mathbf{I}_{n+m}$, where $\mathbf{I}_{n+m}$ is a (n+m)-by-(n+m) identity matrix;
    \item\label{s:c2.1} the columns of $[\mathbf{X}_\mathrm{C}]_{\mathbf{U}}$ are in the range of $\mathbf{W} \in \mathbb{R}^{(n+m)\times n}$ whose columns form a vector basis;
    \item $\mathbf{W} = \augm{\mathbf{W}_\mathrm{s}}{\mathbf{W}_\mathrm{r}}$,  where $\mathbf{W}_\mathrm{s} \in \mathbb{R}^{(n+m)\times p}$ and $\mathbf{W}_\mathrm{r} \in \mathbb{R}^{(n+m)\times q}$ and n = p+q;
    \item\label{s:c2.2} $\mathbf{W}_\mathrm{r}$ is orthogonal to $\mathbf{W}_\mathrm{s}$, i.e. $\inp{\mathbf{W}_\mathrm{s}}{\mathbf{W}_\mathrm{r}} = \mathbf{0}_{p\times q}$, where $\mathbf{0}_{p\times q}$ is a p-by-q zero matrix,
\end{enumerate}
then,
\begin{equation} \label{eq:c2}
    \sum_{k = 1}^{n \wedge m}R_k^2\big({\mathbf{X},\mathbf{Y}}\big) = \sum_{k = 1}^{n \wedge m}\cos^2\big(\varTheta_k({\mathbf{W},[\mathbf{Y}_\mathrm{C}]_{\mathbf{U}}})\big) = \sum_{k = 1}^{p \wedge m}\cos^2\big(\varTheta_k(\mathbf{W}_\mathrm{s},[\mathbf{Y}_\mathrm{C}]_{\mathbf{U}})\big) + \sum_{k = 1}^{q \wedge m}\cos^2\big(\varTheta_k({\mathbf{W}_\mathrm{r},[\mathbf{Y}_\mathrm{C}]_{\mathbf{U}}})\big)
.\end{equation}
\end{theorem}
\emph{\ref{s:c2.b}} to \emph{\ref{s:c2.2}} indicate the assumptions applied in Theorem \ref{c:2}, where the assumption \emph{\ref{s:c2.0}} can be satisfied by the singular value decomposition (SVD).
The proof is given in \ref{ap:c}.
Using the Cosine Superposition Theorem, the following corollary can be obtained.

\begin{corollary}[Maximum Cosine Theorem] \label{t:2}
If,
\begin{enumerate}[label=\textbf{\emph{C\ref{t:2}.\arabic*}},leftmargin=*,align=left]
    \item\label{s:2.b} $\mathbf{X}_\mathrm{C}$ and $\mathbf{Y}_\mathrm{C}$ are the column-centred matrices of $\mathbf{X} \in \mathbb{R}^{N\times n}$ and $\mathbf{Y} \in \mathbb{R}^{N\times m}$;
    \item\label{s:2.1} $[\mathbf{X}_\mathrm{C}]_{\mathbf{U}}$ and $[\mathbf{Y}_\mathrm{C}]_{\mathbf{U}}$ are the coordinate matrices of $\mathbf{X}_\mathrm{C}$ and $\mathbf{Y}_\mathrm{C}$ with respect to the same matrix $\mathbf{U} \in \mathbb{R}^{N\times (n+m)}$ whose columns form an orthonormal basis, where `orthonormal' means $\inp{\mathbf{U}}{\mathbf{U}} = \mathbf{I}_{n+m}$;
    \item\label{s:2.2} the columns of $[\mathbf{X}_\mathrm{C}]_{\mathbf{U}}$ are in the range of $\mathbf{W} \in \mathbb{R}^{(n+m)\times n}$ whose columns form an orthogonal basis, where $\mathbf{W} = (\mathbf{w}_{1}|\mathbf{w}_{2}|\ldots|\mathbf{w}_{n})$;
    \item\label{s:2.3} the columns of $[\mathbf{Y}_\mathrm{C}]_{\mathbf{U}}$ are in the range of $\mathbf{V} \in \mathbb{R}^{(n+m)\times m}$ whose columns form an orthogonal basis, where $\mathbf{V} = (\mathbf{v}_1|\mathbf{v}_{2}|\ldots|\mathbf{v}_m)$;
\end{enumerate}
then,
\begin{equation} \label{eq:ccct}
    \sum_{k = 1}^{n \wedge m}R_k^2({\mathbf{X},\mathbf{Y}}) = \sum_{k = 1}^{n \wedge m}\cos^2\big(\varTheta_k({\mathbf{W},\mathbf{V}})\big) = \sum_{i = 1}^{n}\sum_{j = 1}^{m}\eta_{i,j}
,\end{equation}
where,
\begin{equation} \label{eq:theta}
    \eta_{i,j} = \cos^2\big({\angle(\mathbf{w}_i,\mathbf{v}_j})\big)
.\end{equation}
\end{corollary}
\emph{\ref{s:2.b}} to \emph{\ref{s:2.3}} indicate the assumptions applied in Corollary \ref{t:2}.
It is found that the maximisation in \eqref{eq:defcos} can be decomposed to the orthogonalisation and summation.
On this basis, for the $\eta$-cosine method proposed in this paper, the SSC can be decomposed to $\eta$-cosines \eqref{eq:theta} to speed up its computation.

\section{Canonical-correlation-based greedy search} \label{s:greedy}
Based on the theorems in the last section, a novel feature selection algorithm is proposed. 
Assuming $n$ features with $N$ instances to form the candidate feature matrix $\mathbf{X} \in \mathbb{R}^{N\times n}$ and $m$ variables with $N$ instances to form the target matrix $\mathbf{Y} \in \mathbb{R}^{N\times m}$, the feature selection is the process of selecting $t$ features from $\mathbf{X}$, which are beneficial to build a machine learning model.
For a filter-feature-selection method, a criterion is required to rank the usefulness of features.
In this paper, the ranking criterion is the SSC.
Usually, the target is a vector, i.e. $m = 1$, while $m>1$ may happen, for example, when the target is a categorical variable with seven categories, the target can be dummy encoded to a six-column matrix, where each category is encoded by six binary numbers.
When the target is represented by a vector $\mathbf{y}$, the SSC degenerates to the coefficient of determination, i.e. the squared multiple correlation coefficient $R^2({\mathbf{X},\mathbf{y}})$.

At Iteration $i \in \{0,1,\ldots,t-1\}$, the $(i+1)\textsuperscript{th}$ useful feature $\mathbf{x}_{d} \in \mathbb{R}^{N\times 1}$ from the matrix $\mathbf{X} \in \mathbb{R}^{N\times n}$ is found by greedy search, under retaining the features $\mathbf{X}_\mathrm{s} \in \mathbb{R}^{N\times i}$ selected in the previous iterations.
By adopting the SSC as the ranking criterion, the proposed algorithm finds $\mathbf{x}_{d}$, where, 
\begin{equation} \label{eq:greedy}
    d = \argmax_{j} \sum\limits_{k = 1}^{(i+1)\wedge m}R_k^2\big(\augm{\mathbf{X}_\mathrm{s}}{\mathbf{x}_j},\mathbf{Y}\big)
.\end{equation}

\begin{table}[ht]
  \centering
  \renewcommand{\arraystretch}{2.2}
    \begin{tabular}{ l c l }
    \toprule[2pt]
     Iteration & Criterion & $\mathbf{X}_\mathrm{s}$\\
    \midrule[1pt]
     0 & $R^2({\mathbf{x}_3,\mathbf{Y}}) \geq R^2({\mathbf{x}_j,\mathbf{Y}})$ & $\mathbf{x}_3$ \\
     1 & $\sum\limits_{k = 1}^{2\wedge m}R_k({\augm{\mathbf{x}_3}{\mathbf{x}_5},\mathbf{Y}})\geq \sum\limits_{k = 1}^{2\wedge m}R_k({\augm{\mathbf{x}_3}{\mathbf{x}_j},\mathbf{Y}})$ & $\mathbf{x}_3,\mathbf{x}_5$ \\
     2 & $\sum\limits_{k = 1}^{3\wedge m}R_k({(\mathbf{x}_3|\mathbf{x}_5|\mathbf{x}_1),\mathbf{Y}})\geq \sum\limits_{k = 1}^{3\wedge m}R_k({(\mathbf{x}_3|\mathbf{x}_5|\mathbf{x}_j),\mathbf{Y}})$ & $\mathbf{x}_3,\mathbf{x}_5,\mathbf{x}_1$ \\
    \bottomrule[2pt]
    \end{tabular}
    \caption{An example of selecting three features via greedy search based on the SSC.}
  \label{tbl:greedym}
\end{table}

The canonical correlation coefficients in the SSC can be calculated by the definition given in \eqref{eq:defcc} (called the baseline method), the $h$-correlation in \eqref{eq:t1}, or the $\eta$-cosine in \eqref{eq:ccct}.
Table \ref{tbl:greedym} gives an example of selecting three features by greedy search.
It can be found that as the size of $\mathbf{X}_\mathrm{s}$ increases with the number of iterations, the calculation of canonical correlation coefficients by the baseline method becomes increasingly expensive.
Specifically, for each candidate feature $\mathbf{x}_j$ in Iteration $i$, the computational complexity of the dot product of the column-centred $(\mathbf{X}_\mathrm{s},\mathbf{x}_j)$ required in \eqref{eq:ccc0} is $\mathcal{O}(i^2 N)$, which is increasing with the number of iterations $i$, where $\mathcal{O}$ is the asymptotic upper bound notation \cite{cormen2009introduction}.
However, for the $h$-correlation and $\eta$-cosine methods, the above-described problem can be avoided.

For the $h$-correlation method, according to the Correlation Superposition Theorem, the computation of \eqref{eq:greedy} can be decomposed as follows,
\begin{subequations}
\begin{align}
    \sum\limits_{k = 1}^{(i+1)\wedge m}R_k^2\big(\augm{\mathbf{X}_\mathrm{s}}{\mathbf{x}_j},\mathbf{Y}\big) &= \sum\limits_{k = 1}^{(i+1)\wedge m}R_k^2\big(\augm{\mathbf{W}_\mathrm{s}}{\mathbf{w}_j},\mathbf{Y}\big)\\ 
    &=\sum_{k = 1}^{i \wedge m}R_k^2({\mathbf{W}_\mathrm{s},\mathbf{Y}}) + R^2({\mathbf{w}_j,\mathbf{Y}})\label{eq:decomp}
,\end{align}
\end{subequations}
where the centred columns of $\mathbf{X}_\mathrm{s}$ are in the range of $\mathbf{W}_\mathrm{s} \in \mathbb{R}^{N\times i}$ whose columns are centred and form an orthogonal basis, and the centred $\mathbf{x}_j$ are in the range of $\augm{\mathbf{W}_\mathrm{s}}{\mathbf{w}_j} \in \mathbb{R}^{N\times (i+1)}$ whose columns are centred and form an orthogonal basis.
As the first term on the right-hand side of \eqref{eq:decomp} is the same for the different candidate $\mathbf{x}_j$, only the second term needs to be computed to find the maximum SSC, whose computational complexity is not increasing with the number of iterations $i$.
Furthermore, according to the Maximum Correlation Theorem, if the centred columns of $\mathbf{Y}$ are in the range of $\mathbf{V} \in \mathbb{R}^{N\times m}$, where $\mathbf{V} = (\mathbf{v}_1|\mathbf{v}_{2}|\ldots|\mathbf{v}_m)$, whose columns are centred and form an orthogonal basis, then the second term can be computed by
\begin{equation}\label{eq:r2}
    R^2({\mathbf{w}_j,\mathbf{Y}}) = R^2({\mathbf{w}_j,\mathbf{V}}) = \sum_{k = 1}^{m}r^2(\mathbf{w}_j,\mathbf{v}_{k})
\end{equation}
whose computational complexity is $\mathcal{O}(Nm)$.

According to the Cosine Superposition Theorem, the $\eta$-cosine method can also decompose the computation of \eqref{eq:greedy} in a form similar to \eqref{eq:decomp}.
In addition, as $\mathbf{X}$ and $\mathbf{Y}$ are first transformed into an $(n+m)$-by-$n$ coordinate matrix and an $(n+m)$-by-$m$ coordinate matrix, the computational complexity of \eqref{eq:r2} changes from $\mathcal{O}(Nm)$ to $\mathcal{O}\left((n+m)m\right)$.
Consequently, when $N > m+n$, the $\eta$-cosine method is more efficient than the $h$-correlation method in the long run.
In summary, the computational complexities of the baseline method, the $h$-correlation method, and the $\eta$-cosine method are compared in Table \ref{tbl:bigO}.

\begin{table}[ht]
  \centering
  \renewcommand{\arraystretch}{2}
    \begin{tabular}{ l c l }
    \toprule[2pt]
     Method & $\mathcal{O}$ in Iteration $i$ & Overall $\mathcal{O}$\\
    \midrule[1pt]
     Baseline & $\mathcal{O}(i^2Nn)$ & $\mathcal{O}(t^3Nn)$ \\
     $h$-correlation & $\mathcal{O}(Nnm)$ & $\mathcal{O}(tNnm)$ \\
     $\eta$-cosine & $\mathcal{O}(n^2m)$ & $\mathcal{O}(tn^2m)$ \\
    \bottomrule[2pt]
    \end{tabular}
    \caption{The computational complexity comparison between the baseline method, the $h$-correlation method, and the $\eta$-cosine method, when the feature matrix is $\mathbf{X} \in \mathbb{R}^{N\times n}$, the target matrix is $\mathbf{Y} \in \mathbb{R}^{N\times m}$, $t$ features are selected, and $m \ll t \ll n \ll N$.}
  \label{tbl:bigO}
\end{table}

\HL{Based on the above discussion, the $h$-correlation method and the $\eta$-cosine method can be unified to develop a fast feature selection algorithm, whose pseudocode is given in Algorithm \ref{alg:ffs},} \RV{and the corresponding codes are available in the GitHub repository \footnote{\url{https://github.com/MatthewSZhang/fastcan}} and MATLAB File Exchange \footnote{\url{https://www.mathworks.com/matlabcentral/fileexchange/168911-fastcan}}.}
It is noted that if the $\eta$-cosine method is used, $\mathbf{X}_{\mathrm{C}} \in \mathbb{R}^{N \times n}$ and $\mathbf{Y}_{\mathrm{C}} \in \mathbb{R}^{N \times m}$ should be replaced by $[\mathbf{X}_\mathrm{C}]_{\mathbf{U}} \in \mathbb{R}^{(n+m) \times n}$ and $[\mathbf{Y}_\mathrm{C}]_{\mathbf{U}} \in \mathbb{R}^{(n+m) \times m}$, where the orthonormal matrix $\mathbf{U} \in \mathbb{R}^{N \times (n+m)}$ can be obtained by the SVD of the matrix $\augm{\mathbf{X}_\mathrm{C}}{\mathbf{Y}_\mathrm{C}}$.
Moreover, the orthogonal transformation in Line \ref{alg:svd}, Line \ref{alg:orth} and Line \ref{alg:mgs} can be realised by various orthogonalisation methods, such as the classical Gram-Schmidt process, the modified Gram-Schmidt process, and the Householder transformation.
However, for the sake of computation speed, the modified Gram-Schmidt process should be used in Line \ref{alg:mgs}.
\HL{Meanwhile, the modified Gram-Schmidt process suffers from a large rounding error when it is applied to a set of vectors which are almost linearly dependent \cite{golub2013matrix}, and this property is used in Line \ref{alg:ldc} of Algorithm 1, to skip the redundant features.}

\begin{algorithm}
\caption{Pseudocode for the fast feature selection algorithm based on $h$-correlation and $\eta$-cosine}\label{alg:ffs}
\begin{algorithmic}[1]
\Require{$\mathbf{X} \in \mathbb{R}^{N \times n}$ \Comment{Candidate feature matrix}\\%
$\mathbf{Y} \in \mathbb{R}^{N \times m}$ \Comment{Target matrix}\\%
$t \in \mathbb{N}$ \Comment{Number of features to be selected}\\%
$\epsilon \in \mathbb{R}$  \Comment{Threshold for linear dependence checks}
}
\Ensure $\mathbf{d}\in \mathbb{R}^{1\times t}$, where $\mathbf{d} = (d_{1}, d_{2}, \ldots, d_{t})$ \Comment{Indices of the selected features}
\State $\mathbf{X}_{\mathrm{C}} \gets \mathbf{X} - \overline{\mathbf{X}}$ \Comment{Centre $\mathbf{X}$ to $\mathbf{X}_{\mathrm{C}}$, where $\overline{\mathbf{X}}$ is the matrix formed by the mean of each column of $\mathbf{X}$}
\State $\mathbf{Y}_{\mathrm{C}} \gets \mathbf{Y} - \overline{\mathbf{Y}}$ \Comment{Centre $\mathbf{Y}$ to $\mathbf{Y}_{\mathrm{C}}$, where $\overline{\mathbf{Y}}$ is the matrix formed by the mean of each column of $\mathbf{Y}$}
\If{\RV{use $h$-correlation}}
    \State $\mathbf{A} \gets \mathbf{X}_{\mathrm{C}}$
    \State $\mathbf{B} \gets \mathbf{Y}_{\mathrm{C}}$
\EndIf
\If{use $\eta$-cosine}
    \State $\mathbf{U}\mathbf{S}\mathbf{V}_\mathrm{h}^\top  \gets \svd\left({\augm{\mathbf{X}_{\mathrm{C}}}{\mathbf{Y}_{\mathrm{C}}}}\right)$ \Comment{Singular value decomposition}
    \label{alg:svd}
    \State $\augm{[\mathbf{X}_{\mathrm{C}}]_\mathbf{U}}{[\mathbf{Y}_{\mathrm{C}}]_\mathbf{U}} \gets \mathbf{S}\mathbf{V}_\mathrm{h}^\top$ \Comment{Due to Proposition \ref{p:1}}
    \State $\mathbf{A} \gets [\mathbf{X}_{\mathrm{C}}]_\mathbf{U}$
    \State $\mathbf{B} \gets [\mathbf{Y}_{\mathrm{C}}]_\mathbf{U}$
\EndIf
\State $\mu_i \gets 1,\; i=1,\ldots,n$ \Comment{Initialise the mask vector for valid candidate features}
\State $R_i^2 \gets 0,\; i=1,\ldots,n$ \Comment{Initialise the feature ranking score vector}
\State $d_i \gets 0,\; i=1,\ldots,t$ \Comment{Initialise the output indices vector}
\For{$i \gets 1$ to $t$}
    \If{$i = 1$} \Comment{Preprocess $\mathbf{A}$ and $\mathbf{B}$}
        \State $\mathbf{V} \gets \orth\left({\mathbf{B}}\right)$ \Comment{Orthonormal transformation from $\mathbf{B}$ to $\mathbf{V}$}
        \label{alg:orth}
        \State $\mathbf{W} \gets \normalise\left({\mathbf{A}}\right)$, where $\mathbf{W}=(\mathbf{w}_{1}|\mathbf{w}_{2}|\ldots|\mathbf{w}_{n})$ \Comment{Normalising the columns of $\mathbf{A}$ to unit variances}
    \Else
        \State $\mu_d \gets 0$ \Comment{Mask the selected feature}
        \State $R^2_{d} \gets 0$ \Comment{Set the ranking score of the selected feature to 0 to avoid selecting that feature again}
        \For{$j \gets 1$ to $n$} \Comment{Make candidate features orthogonal to the selected features}
            \If{$\mu_j = 1$}
                \State $\mathbf{w}_{j} \gets \mathbf{w}_{j} - \mathbf{w}_{d}\mathbf{w}_{d}^\top\mathbf{w}_{j}$ \Comment{Make $\mathbf{w}_j$ orthogonal to $\mathbf{w}_{d}$ via the modified Gram-Schmidt process}
                \label{alg:mgs}
                \State $\mathbf{w}_{j} \gets \normalise\left({\mathbf{w}_{j}}\right)$ \Comment{Normalising $\mathbf{w}_{j}$ to unit variance}
                \State $g \gets \mathbf{w}_{j}^\top\mathbf{w}_{d}$ \Comment{Compute rounding error of orthogonalisation}
                \If{$\lvert g \rvert>\epsilon$} \Comment{\HL{Linear dependence check}, where $\lvert \cdot \rvert$ denotes an absolute value}
                \label{alg:ldc}
                    \State $\mu_j \gets 0$ \Comment{Mask the nearly linearly dependent feature}
                    \State \RV{$R^2_{j} \gets 0$} \Comment{Avoid selecting the nearly linearly dependent feature}
                \EndIf
            \EndIf
        \EndFor
    \EndIf
    \For{$j \gets 1$ to $n$}
        \If{$\mu_j = 1$}
            \State \RV{$\mathbf{r} \gets \mathbf{w}_{j}^\top \mathbf{V}$} \Comment{Due to \eqref{eq:r2}, compute $\mathbf{r} \in \mathbb{R}^{1\times m}$, which is composed of $r(\mathbf{w}_j,\mathbf{v}_{k}),\; k=1,\ldots,m$}
            \State $R_j^2 \gets \mathbf{r} \mathbf{r}^\top$ \Comment{Compute the sum of squared $r(\mathbf{w}_j,\mathbf{v}_{k})$}
        \EndIf
    \EndFor
    \State $d \gets \argmax_{j} R_j^2$ \Comment{Find the index of the feature with the largest ranking score}
    \State $d_i \gets d$

\EndFor
\end{algorithmic}
\end{algorithm}

\section{Empirical studies}\label{s:case studies}
In this section, the elapsed time of the $h$-correlation and $\eta$-cosine methods is first compared with the baseline method to visually illustrate the computational speed improvement of the two fast methods. 
\RV{Furthermore, by comparing the elapsed time of the $h$-correlation method with that of the proposed $\eta$-cosine method, the applicable scenario and speed advantage of the $\eta$-cosine method are more intuitively demonstrated.}
Secondly, eight real datasets are used to comprehensively compare the SSC with seven other feature ranking criteria \cite{brown2012conditional} on their effectiveness in selecting useful features for general classification and regression tasks. \RV{The corresponding feature selection methods used here are all filter methods, and all features are selected in a greedy manner.}
\RV{The comparisons here help to justify studying feature selection methods based on the SSC.} 

\RV{Finally, three case studies based on two experimental datasets and one simulated dataset are used to show the application scenarios and performance of the proposed algorithm in the SHM field.}
Using the data from a Gnat trainer, the proposed fast feature selection algorithm is compared with eight state-of-the-art feature selection methods in \texttt{scikit-learn} \cite{scikit-learn}, in terms of selecting damage-sensitive features.
\RV{These eight methods cover three types of feature selection methods, namely filter, wrapper, and embedded methods. 
This case study aims to further illustrate the characteristics of SSC-based feature selection and its advantages in SHM.}
\RV{The second case study on a glider wing aims to check the performance of the proposed algorithm under environmental variations for detecting different damage cases.}
\RV{The third case study is used to demonstrate the feasibility and performance of the proposed SSC-based algorithm deployed in an edge computing device. 
Some comparisons between the proposed algorithm and the methods in \texttt{scikit-learn} are also made in the last two case studies.}

\subsection{Comparison of elapsed time} \label{ss:time}
\begin{figure}[ht]
\centering
\includegraphics[width=0.55\linewidth]{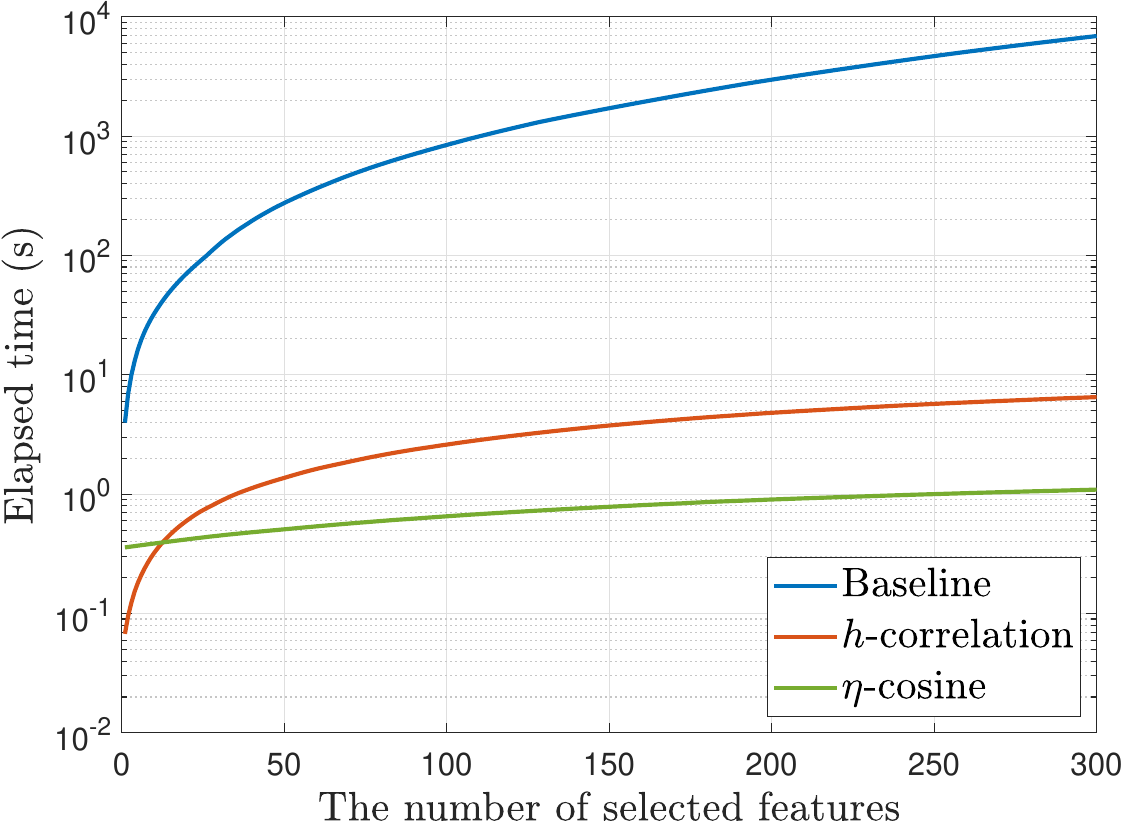}
\caption{The elapsed time of the definition-, $h$-correlation- and $\eta$-cosine-based feature selection methods.}
\label{fig:time}
\end{figure}

A synthetic dataset is generated to compare the running time of the $h$-correlation and $\eta$-cosine methods on feature selection tasks. The built-in function {\tt canoncorr} of MATLAB$^\circledR$ is adopted as the baseline method to compute this criterion as well.
The feature and target matrices $\mathbf{X} \in \mathbb{R}^{N \times n}$ and $\mathbf{Y} \in \mathbb{R}^{N \times m}$ are generated from random numbers uniformly distributed between 0 to 1.
The number of instances is 5000, and the number of variables in $\mathbf{X}$ and $\mathbf{Y}$ are 700 and 50, respectively, i.e. $N = 5000$, $n = 700$, $m = 50$.
All three methods are used to greedily select the features according to the SSC.
The results are shown in Figure \ref{fig:time}. \RV{Features here refer to the variables in X.}

It can be seen that the computational speed of the baseline method is much slower than the other two methods, even at the first iteration.
According to the analysis in Section \ref{s:greedy}, when $N > n+m$, the $\eta$-cosine method is generally faster than the $h$-correlation method.
However, Figure \ref{fig:time} shows that the $\eta$-cosine method is slower than the $h$-correlation method until the selection of the $13\textsuperscript{th}$ feature.
This phenomenon appears because, for the $\eta$-cosine method, before the first iteration, the orthonormal basis $\mathbf{U}$ needs to be found by the SVD at Line \ref{alg:svd} in Algorithm \ref{alg:ffs}, which dominates the computational complexity of selecting the first several features.
However, since the SVD only needs to be done once, the $\eta$-cosine method is faster than the $h$-correlation method in the long run.

\subsection{Applications to the open-access datasets}
The criterion used in the proposed algorithm, i.e. SSC, is compared with seven other feature ranking criteria, which are minimal-Redundancy-Maximal-Relevance (mRMR), Mutual Information Maximisation (MIM), Joint Mutual Information (JMI), Conditional Mutual Information Maximisation (CMIM), Conditional Infomax Feature Extraction (CIFE), Interaction Capping (ICAP), and Double Input Symmetrical Relevance (DISR) \cite{brown2012conditional}. 

Three UCI datasets \cite{Dua:2019} and the MNIST dataset \cite{lecun1998gradient}, which are summarised in Table \ref{tbl:class}, are used to compare the SSC with the seven criteria in selecting features for classification tasks.
The \textit{Lymph} dataset \cite{michalski1986multi} aims to classify lymphography information into two classes, i.e. metastases and malign lymph, by 18 medical diagnostic attributes \RV{as features}.
The CNAE dataset \cite{ciarelli2009agglomeration} is designed to classify 1080 documents of business descriptions of Brazilian companies into nine economic activities based on the occurrence frequency of 856 words in the documents \RV{as features}.
The Mfeat dataset \cite{van1998handwritten} aims to classify the handwritten numerals (i.e. 0 to 9) in a collection of Dutch utility maps by 649 features extracted from the raw images.
The MNIST dataset \cite{lecun1998gradient} is designed to classify the handwritten digits using the grey level of 784 pixels \RV{as features}.

\begin{table}[ht]
  \centering
  \renewcommand{\arraystretch}{1.5}
    \begin{tabular}{ l c c c c }
    \toprule[2pt]
      & Lymph & CNAE & Mfeat & MNIST \\
    \midrule[1pt]
    Data Type & \makecell{Categorical\\\& Discrete} & Discrete & Continuous & Discrete \\
    No. of Instances & 142 & 1080 & 2000 & 700000 \\
    No. of Features & 18 & 856 & 649 & 784 \\
    No. of Classes & 2 & 9 & 10 & 10 \\
    Classifier & SVM & LDA & SVM & LDA \\
    \bottomrule[2pt]
    \end{tabular}
    \caption{Four open-access datasets for classification tasks.}
  \label{tbl:class}
\end{table}

\begin{figure}[ht]
\centering
\begin{subfigure}[b]{0.35\textwidth}
\centering
\includegraphics[width=1\linewidth]{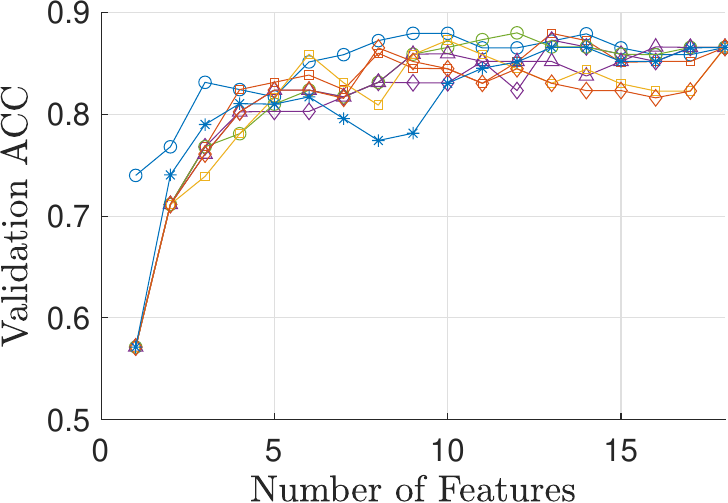}
\caption{Lymph}
\label{fig:lymph}
\end{subfigure}
\hspace{1cm}
\begin{subfigure}[b]{0.35\textwidth}
\centering
\includegraphics[width=1\linewidth]{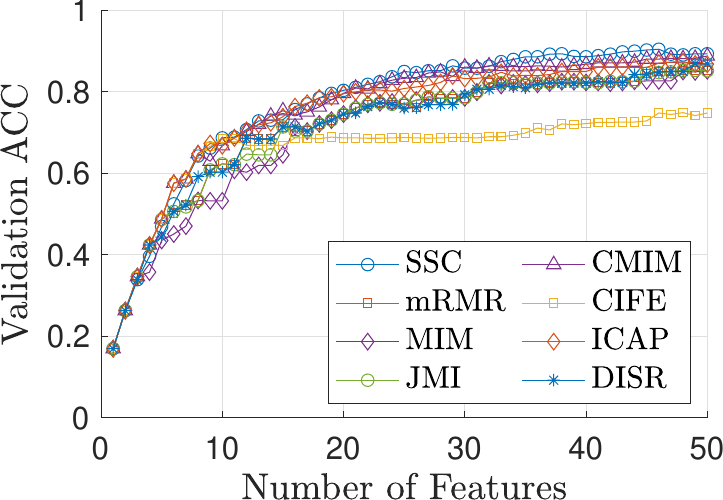}
\caption{CNAE}
\label{fig:CNAE}
\end{subfigure}

\begin{subfigure}[b]{0.35\textwidth}
\centering
\includegraphics[width=1\linewidth]{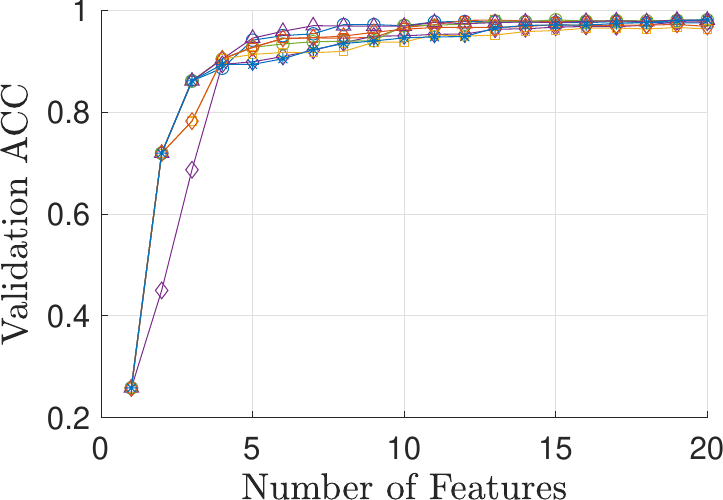}
\caption{Mfeat}
\label{fig:mfeat}
\end{subfigure}
\hspace{1cm}
\begin{subfigure}[b]{0.35\textwidth}
\centering
\includegraphics[width=1\linewidth]{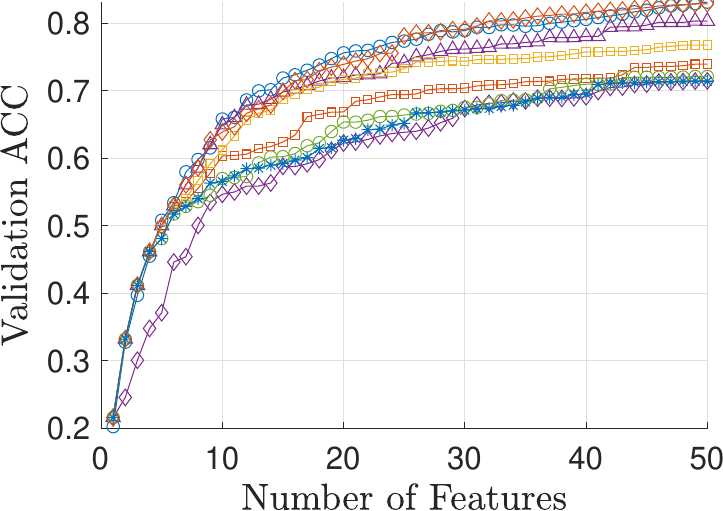}
\caption{MNIST}
\label{fig:MNIST}
\end{subfigure}

\begin{subfigure}[b]{0.9\textwidth}
\centering
\includegraphics[width=1\linewidth]{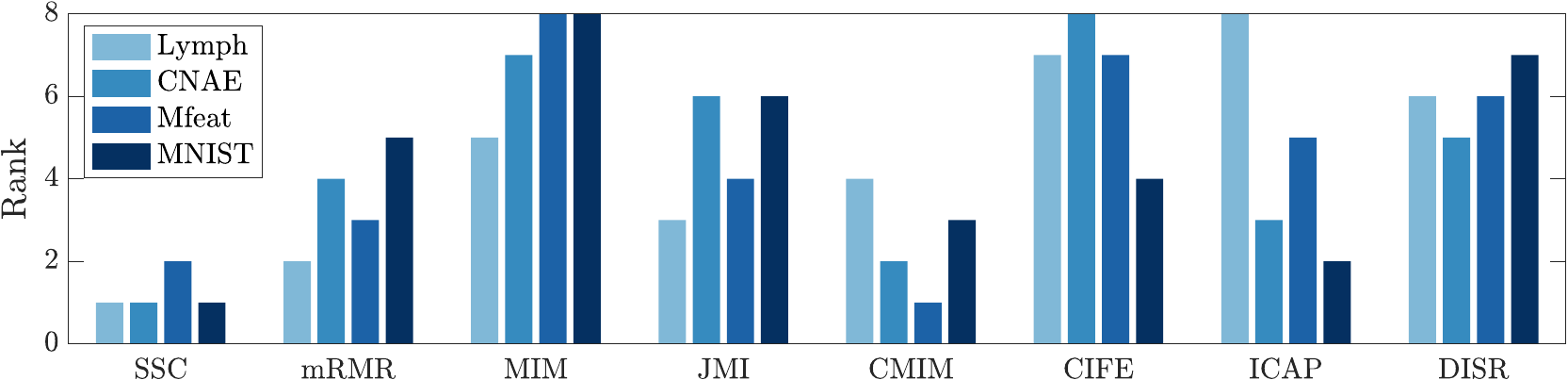}
\caption{Rank of criteria in the four datasets according to their averaged ratios between the validation ACC and the number of selected features.}
\label{fig:rank_class}
\end{subfigure}

\caption{Comparison of the averaged ACC results between different feature ranking criteria in classification tasks.}
\label{fig:class}
\end{figure}

For the SSC, the categorical features are assigned values by ordinal encoding, and the target variable is encoded by dummy encoding. 
For the other seven criteria, the continuous features are discretised into five equal-width bins.
Given the selected features, a Support Vector Machine (SVM) or a Linear Discriminant Analysis (LDA) model is adopted for classification tasks, where the continuous features are standardised to z-scores.
Ten-fold cross-validation is applied.
The performance of the feature selection criteria is evaluated by the averaged classification accuracy (ACC) on the ten validation datasets.
The results are shown in Figure \ref{fig:class}.
Generally, the SSC gives competitive results, whether the number of the selected features is small or large.
When the ACC results are averaged by the number of selected features, the SSC achieves the best results in three of the four datasets, as shown in Figure \ref{fig:rank_class}.

Four additional UCI datasets \cite{Dua:2019}, which are summarised in Table \ref{tbl:regression}, are used to compare the SSC with the seven other criteria in selecting features for regression tasks.
The \textit{Student} dataset \cite{cortez2008using} aims to predict the final grade (from 0 to 20) of the Portuguese class in secondary education of two Portuguese schools by demographic, social and school-related features.
The \textit{Parkinson} dataset \cite{sakar2013collection} is designed to predict the Unified Parkinson Disease Rating Scale (UPDRS) scores by features extracted from multiple types of sound recordings, including sustained vowels, numbers, words and short sentences.
The \textit{Conductor} dataset \cite{hamidieh2018data} aims to predict the critical temperature of a superconductor from features extracted from the superconductor’s chemical formula.
The \textit{Energy} dataset \cite{candanedo2017data} is designed to predict the energy usage of appliances during operation from environmental parameters \RV{as features}, where the original 24 features are expanded to 2924 features using 1\textsuperscript{st} to 3\textsuperscript{rd}-degree polynomial basis functions.

\begin{figure}[htbp!]
\centering
\begin{subfigure}[b]{0.35\textwidth}
\centering
\includegraphics[width=1\linewidth]{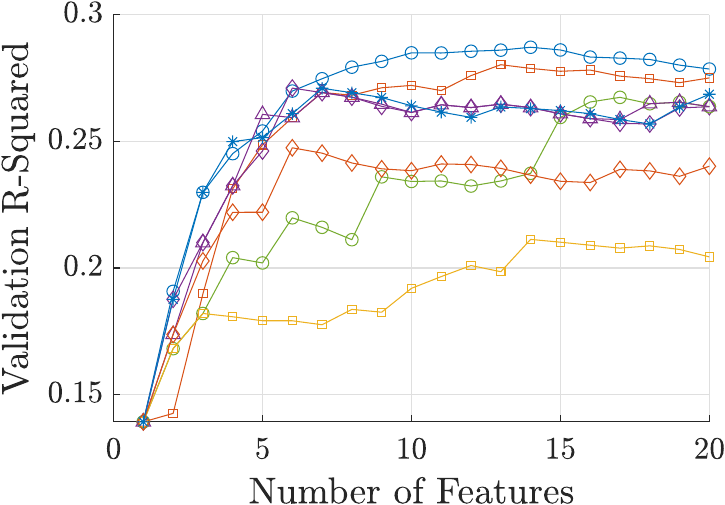}
\caption{Student}
\label{fig:studentPerformance}
\end{subfigure}
\hspace{1cm}
\begin{subfigure}[b]{0.35\textwidth}
\centering
\includegraphics[width=1\linewidth]{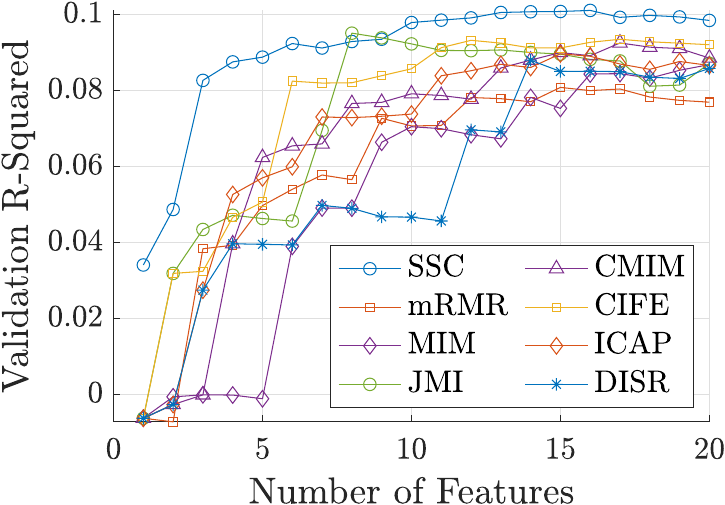}
\caption{Parkinson}
\label{fig:Parkinson}
\end{subfigure}

\begin{subfigure}[b]{0.35\textwidth}
\centering
\includegraphics[width=1\linewidth]{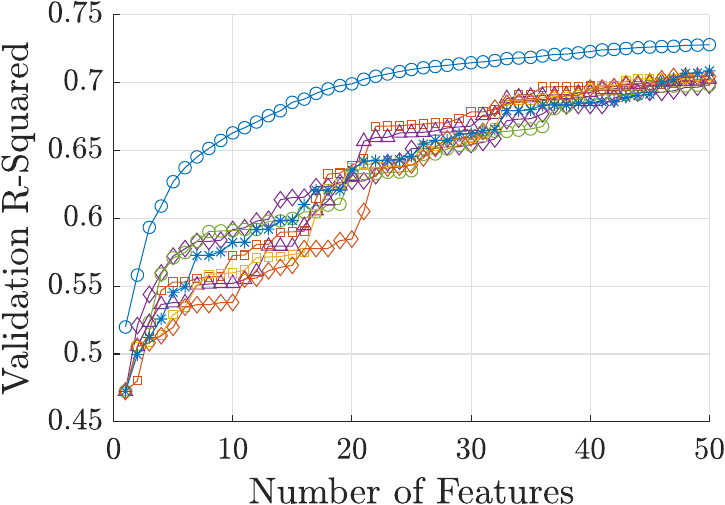}
\caption{Conductor}
\label{fig:Conductor}
\end{subfigure}
\hspace{1cm}
\begin{subfigure}[b]{0.35\textwidth}
\centering
\includegraphics[width=1\linewidth]{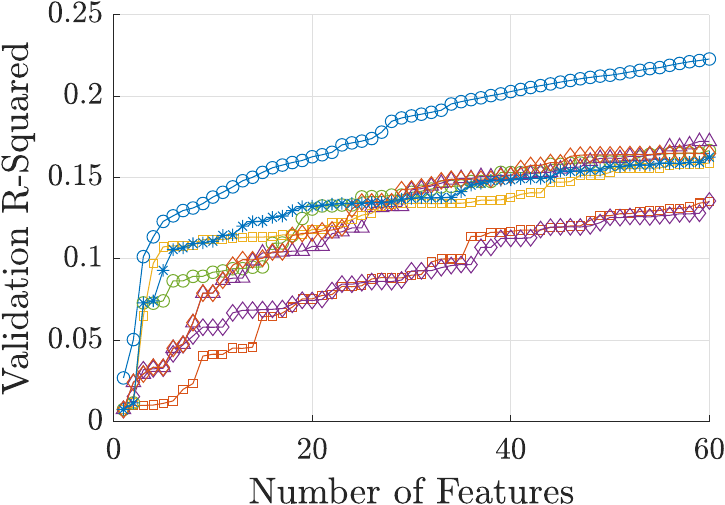}
\caption{Energy}
\label{fig:Energy}
\end{subfigure}

\begin{subfigure}[b]{0.9\textwidth}
\centering
\includegraphics[width=1\linewidth]{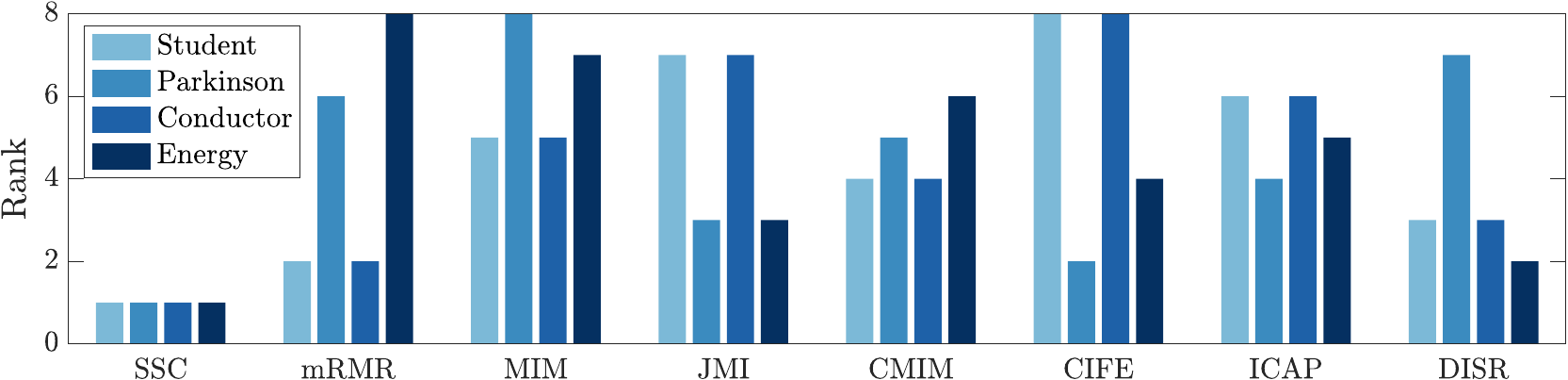}
\caption{Rank of criteria in the four datasets according to their averaged ratios between the validation \RV{R-squared} and the number of selected features.}
\label{fig:rank_reg}
\end{subfigure}

\caption{Comparison of the averaged \RV{R-squared} results between different feature ranking criteria in regression tasks.}
\label{fig:regression}
\end{figure}

Similar to the classification tasks, for the SSC, the categorical features are assigned ordinal values.
For the seven other criteria, the continuous features of the Energy dataset are discretised into ten equal-width bins, while the others are into five equal-width bins.
Given the selected features, a Linear Regression (LR) model is applied for regression tasks, where the continuous features are standardised to z-scores.
Ten-fold cross-validation is applied.
The performance of the feature selection criteria is evaluated by the averaged \RV{R-squared} on the ten validation datasets.
The results are shown in Figure \ref{fig:regression}.
The SSC has an obvious advantage over the other criteria.
It should be noted that since all four tasks are single-output regressions, the SSC degenerates to the coefficient of determination, which monotonically \RV{increases} with the \RV{R-squared} of the LR model \cite{glantz2016primer}.
Therefore, in these four cases, the SSC-based filter method is equivalent to the LR-based wrapper method.

\begin{table}[htbp!]
  \centering
  \renewcommand{\arraystretch}{1.5}
    \begin{tabular}{ l c c c c }
    \toprule[2pt]
      & Student & Parkinson & Conductor & Energy\\
    \midrule[1pt]
    Data Type & \makecell{Categorical\\\& Discrete} & Continuous & Continuous & Continuous \\
    No. of Instances & 649 & 1040 & 21263 & 19735 \\
    No. of Features & 30 & 26 & 81 & 2924 \\
    \bottomrule[2pt]
    \end{tabular}
    \caption{Four open-access datasets for regression tasks.}
  \label{tbl:regression}
\end{table}

\subsection{A case study on data from a Gnat trainer aircraft}
\begin{figure}[ht]
    \centering
    \includegraphics[width=0.53\linewidth]{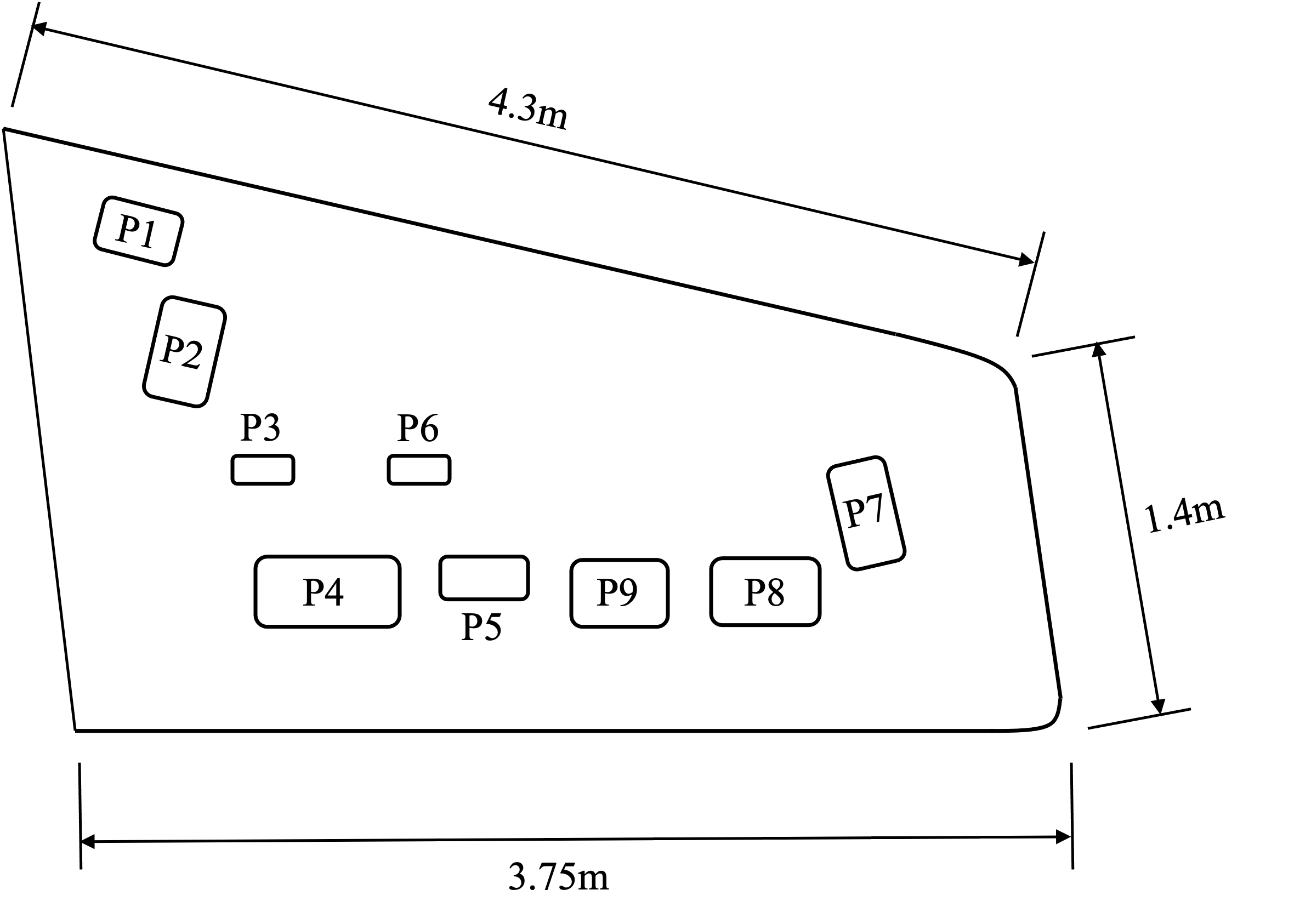}
    \caption{The schematic of the starboard wing and nine inspection panels.}
    \label{fig:schematic_wing}
\end{figure}

An engineering dataset collected from the starboard wing of a Gnat trainer aircraft is used to demonstrate the effectiveness of the proposed algorithm in damage-sensitive feature selection.
The schematic of the starboard wing is shown in Figure \ref{fig:schematic_wing}.
The nine damage scenarios were simulated by sequentially removing nine inspection panels, whose sizes and distribution can also be seen in Figure \ref{fig:schematic_wing}.
The 44 candidate features in total were extracted from measured transmissibilities, which are the ratios of the acceleration response spectra and the reference acceleration response spectra \cite{worden2008genetic,gardner2022application}. 
The data contains 1782 instances and each damage scenario has 198 instances.
Nine damage-sensitive features will be selected in this case study to train an SVM classifier for damage prediction.
For a more detailed description of the test set-up, data collection process and feature extraction process, please refer to \cite{manson2003experimental}.

In the damage detection task, the proposed algorithm is compared with eight feature selection methods in scikit-learn \cite{scikit-learn}, which are highly optimised and popular among machine learning practitioners. Note that \texttt{scikit-learn} is a popular machine learning library for Python programming language.
\HL{The eight feature selection methods from scikit-learn (version 1.3.0) correspond to four different algorithms, namely the Select From a Model (SFM) algorithm, the Recursive Feature Elimination (RFE) algorithm, the Sequential Feature Selection (SFS) algorithm, and Select K Best (SKB) algorithm.}
The SFM algorithm, belonging to embedded feature selection, ranks features based on their weights in a given classifier.
Similarly, the RFE algorithm, belonging to embedded feature selection, recursively trains a given classifier and eliminates the least important features based on the weights of features \cite{guyon2002gene}.
The SFS algorithm, belonging to wrapper feature selection, adds (forward selection) or removes (backward selection) features based on the cross-validation performance of a given classifier.
The SKB algorithm, belonging to filter feature selection, selects features according to the feature scores defined by score functions, e.g., Pearson's correlation, mutual information, etc.
Specifically, the methods in this comparison are 
\begin{itemize}
    \item \texttt{SSC\_eta}, which is the fast SSC-based feature selection algorithm accelerated with $\eta$-cosine,
    \item \texttt{SSC\_h}, which is the fast SSC-based feature selection algorithm accelerated with $h$-correlation,
    \item \texttt{SFM\_lsvc}, which is the SFM algorithm with a linear support vector classifier,
    \item \texttt{SFM\_rfc}, which is the SFM algorithm with a random forest classifier,
    \item \texttt{RFE\_lsvc}, which is the RFE algorithm with a linear support vector classifier,
    \item \texttt{RFE\_rfc}, which is the RFE algorithm with a random forest classifier,
    \item \texttt{SFS\_lsvc}, which is the forward SFS algorithm with a linear support vector classifier,
    \item \texttt{SFS\_rfc}, which is the forward SFS algorithm with a random forest classifier,
    \item \texttt{SKB\_clf}, which is the SKB algorithm ranking features with ANOVA (analysis of variance) F-statistic, and
    \item \texttt{SKB\_mic}, which is the SKB algorithm ranking features with mutual information.
\end{itemize}

Both the computational speed and classification performance of feature selection methods are investigated.
\HL{It should be noted that as the $h$-correlation and $\eta$-cosine method use the same feature ranking criterion, i.e. SSC, the features selected by the $h$-correlation method are the same as those selected by the $\eta$-cosine method.
Therefore, the $h$-correlation and $\eta$-cosine method have no difference in classification performance.}
For the SSC-based feature selection algorithm, both dummy encoding (denoted as SSC(d)) and ordinal encoding (denoted as SSC(o)) are used to encode the labels.
As the other methods only accept a vector as the target, the labels are encoded into a vector by ordinal encoding.
The average elapsed time of 100 tests for each algorithm on the whole dataset is used to compare their computational speed.
The classification accuracy of SVM models on ten validation datasets, which are from the stratified ten-fold cross-validation, is used to compare the classification performance of each feature selection method.
The training dataset is used for feature selection, hyperparameter tuning, and model training. 
The hyperparameters of the SVM are tuned by grid search with five-fold cross-validation over kernel type and $\ell^2$ regularisation parameter.
The kernel type is selected from linear and RBF (i.e. radial basis function), while the regularisation parameter is selected from 0.1, 1, and 10.

\begin{figure}[ht]
    \centering
    \includegraphics[width=0.85\linewidth]{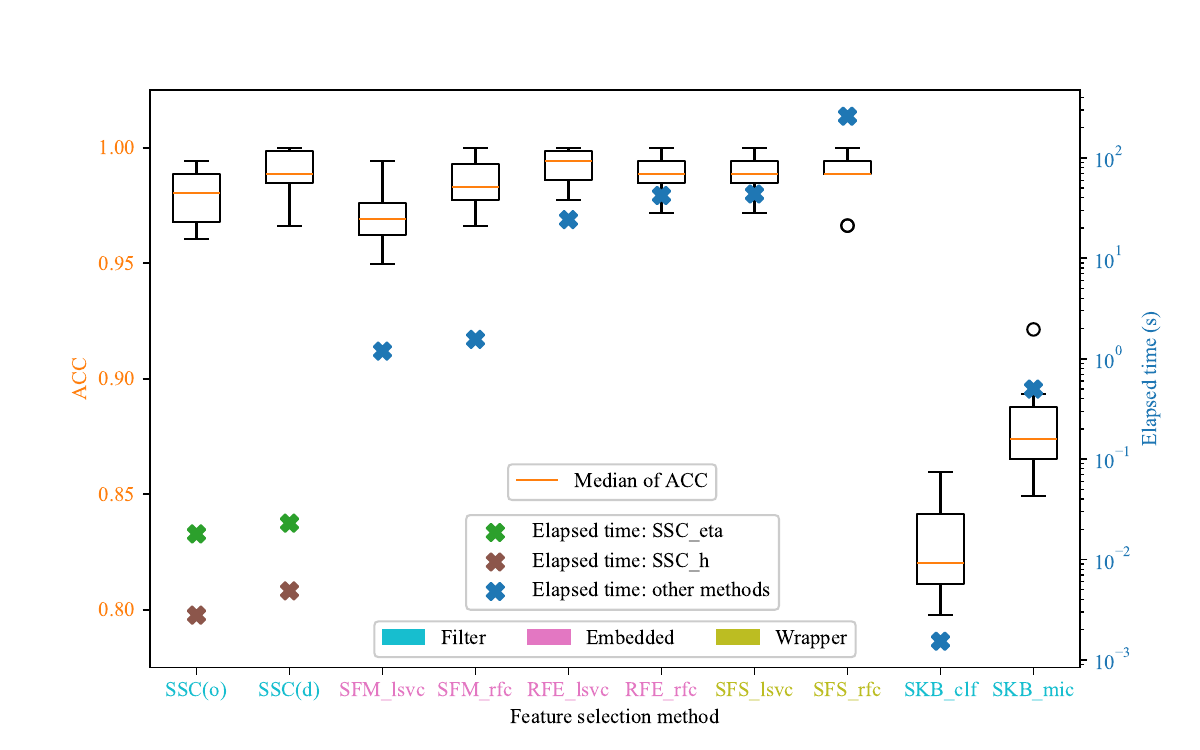}
    \caption{Comparison between the proposed SSC-based algorithm and the eight feature selection methods in scikit-learn on classification accuracy and computational speed, where SSC(o) represents the SSC-based algorithm with ordinal encoding and SSC(d) represents the SSC-based algorithm with dummy encoding.}
    \label{fig:SSCvsWrapper}
\end{figure}

In Figure \ref{fig:SSCvsWrapper}, the classification results are shown by the box plot, while the computational speed results are shown by the scatter plot. 
The orange lines in the boxes are medians of the classification accuracy (ACC) of ten validation datasets.
The boxes indicate the first quartile (Q1) and the third quartile (Q3).
The whiskers are the vertical lines extending to the most extreme excluding outlier data points.
The caps are the horizontal lines at the ends of the whiskers, which are Q1 - 1.5*IQR and Q3 + 1.5*IQR, where IQR is the interquartile range, i.e. Q3-Q1.
The fliers are points representing data that extend beyond the whiskers.

The classification results show that the filter methods generally have worse classification performance but higher computational speed than the embedded methods and the wrapper methods.
However, the proposed filter method not only has a higher computational speed but also has a similar classification performance as the embedded and wrapper methods. 
\RV{One possible reason for this phenomenon is that the interaction and redundancy of feature vectors are considered in the proposed method, which is illustrated by the concrete examples in \ref{ap:d}.}
It is also found that although SSC(d) is slower than SSC(o) but gives better classification performance, which may be due to that ordinal encoding assigns ordered values to qualitative values in the target while dummy encoding maintains the categorical property of the target.
It should be noted that the $h$-correlation method is faster than the $\eta$-cosine method here because the SVD in the $\eta$-cosine method dominates the computational complexity when the number of features to be selected is small (i.e. nine in this case). \RV{This phenomenon} has been analysed in Section \ref{ss:time}.

\subsection{\RV{A case study on data from a glider wing in different environments}}
\begin{figure}[ht]
\centering
\includegraphics[width=0.58\linewidth]{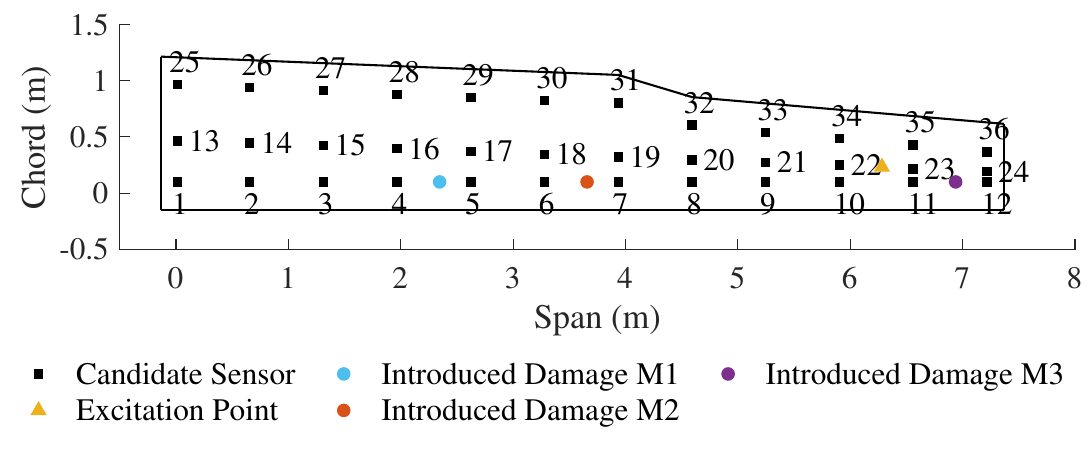}
\caption{Labelled positions of significant points on the glider wing.} \label{fig:glider_sensors}
\end{figure}

\RV{Another engineering dataset collected from a full wing of a glider aircraft under different temperatures is used to illustrate the performance of the proposed algorithm for damage detection tasks under environmental variations. 
The glider wing was tested in a chamber where the ambient temperature around the wing could be controlled as needed.
The dataset contains the 36 features (the square of the Mahalanobis distance) extracted from the frequency response function collected from 36 piezoelectric accelerometers mounted on the wing. One feature is extracted from each sensor. The adopted frequency range was 10 Hz to 50 Hz. 
Figure \ref{fig:glider_sensors} is a schematic showing the locations of 36 sensors.}

\RV{Data for the normal condition, M1 damage (D1) condition, M2 damage (D2) condition, and M3 damage (D3) condition were collected at six different temperatures, i.e. 0, 5, 10, 15, 20 and 25 $^{\circ}$C. Damage was simulated by adding a mass block (60g) onto the structure at three locations, as shown in Figure \ref{fig:glider_sensors}.
Due to the limitation of experimental time, the data for the D3 case at 20 and 25 $^{\circ}$C are absent. At each temperature, there are 50 samples for the normal condition and 25 for each damage condition. 
The normal, D1, D2, and D3 cases are encoded as zero, one, two, and three, respectively, as the target variable.
For more details on the experimental setups, data collection and feature extraction, please refer to \cite{wang2023improving}.}

\begin{figure}[htbp!]
\centering
\begin{subfigure}[b]{0.46\textwidth}
\centering
\includegraphics[width=1\linewidth]{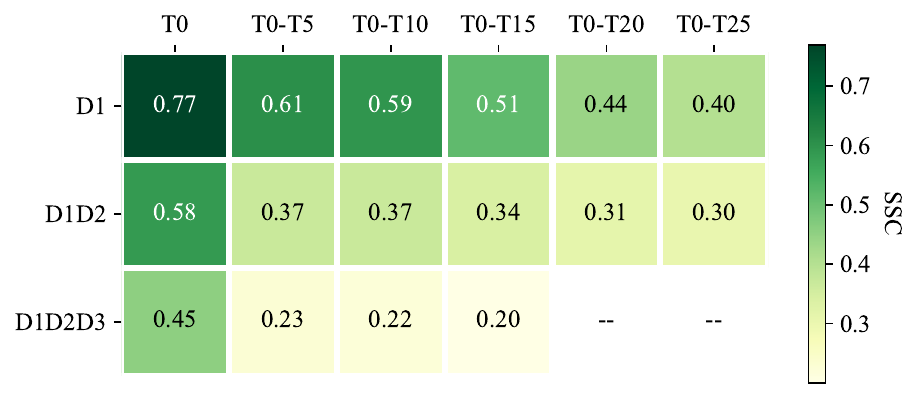}
\caption{Ranking scores}
\label{fig:glider_ssc_score}
\end{subfigure}
\hspace{0.3cm}
\begin{subfigure}[b]{0.46\textwidth}
\centering
\includegraphics[width=1\linewidth]{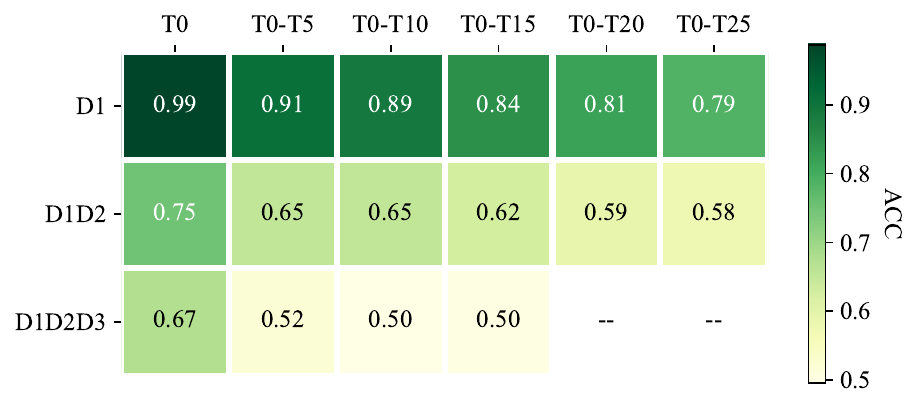}
\caption{Classification accuracy}
\label{fig:glider_ssc_acc}
\end{subfigure}

\caption{Feature ranking scores and classification accuracy corresponding to the features selected by the proposed SSC-based algorithm. D1 means the data collected under the D1 condition and normal condition are used, D1D2 means the data collected under the D1 condition, D2 condition and normal condition are used, and D1D2D3 means the data collected under all four conditions are used. T0 means the data collected at 0 $^{\circ}$C are used, and T0-TX means the data collected from 0 to X $^{\circ}$C at intervals of 5 $^{\circ}$C are used.}
\label{fig:glider_class_ssc}
\end{figure}

\begin{figure}[ht]
\centering
\begin{subfigure}[b]{0.46\textwidth}
\centering
\includegraphics[width=1\linewidth]{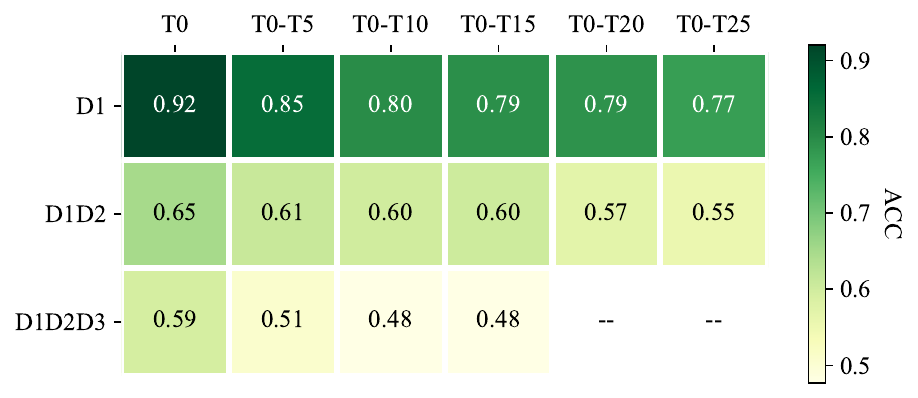}
\caption{\texttt{SKB\_clf}}
\label{fig:glider_skb_mic}
\end{subfigure}
\hspace{0.3cm}
\begin{subfigure}[b]{0.45\textwidth}
\centering
\includegraphics[width=1\linewidth]{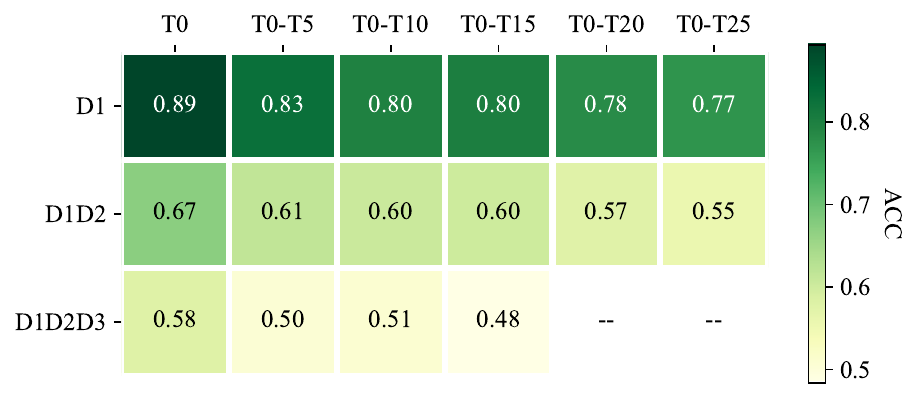}
\caption{\texttt{SKB\_mic}}
\label{fig:glider_skb_clf}
\end{subfigure}

\caption{Classification accuracy corresponding to the features selected by the SKB-based algorithm. SKB\_clf refers to the SKB algorithm ranking features with ANOVA F-statistic. SKB\_mic refers to the SKB algorithm ranking features with mutual information.}
\label{fig:glider_class_others}
\end{figure}

\RV{The feature selection task is to select 12 damage-sensitive features from 36 candidates. Then, the selected features are given to an LDA classifier to demonstrate the damage detection effectiveness. 
By combining data for different temperatures and health conditions, 16 cases were obtained here to consider damage detection tasks of various difficulty levels. These 16 cases can be seen in Figure \ref{fig:glider_class_ssc} and \ref{fig:glider_class_others}. 
Note that temperature information is not used here to deal with the confounding effect caused by temperature variations, because the performance of different filter methods under different degrees of data contamination is expected to be presented and compared.} 

\RV{By comparing the feature ranking scores and classification accuracy corresponding to the features selected by the proposed algorithm in Figure \ref{fig:glider_class_ssc}, it can be found that the changing trends of the two values for the 16 cases are consistent. 
This consistency reflects, to a certain extent, the rationality of using the SSC as the evaluation criterion to select useful features for damage detection tasks.}

\RV{From Figures \ref{fig:glider_ssc_acc}, \ref{fig:glider_skb_clf} and \ref{fig:glider_skb_mic}, it is not difficult to find that whether for the identification of single damage or multiple damages, when data collected at more temperatures are fused, the accuracy of damage detection will deteriorate.
Nevertheless, overall, the SSC-based algorithm has better damage detection performance than the SKB-based algorithm. 
One possible reason for this phenomenon is that the SSC-based algorithm can better consider the redundant information between features.}

\RV{According to the rule of thumb, features from closer sensors share more similar information, so the more features are selected from neighbouring sensors, the worse the corresponding feature selection method handles the redundant features.
To quantitatively measure the redundancy of information in the selected features, the number of neighbouring features contained in the selected features of the three methods is calculated here.
The features provided by two sensors that are horizontally or vertically adjacent in Figure \ref{fig:glider_sensors} are counted as neighbouring features. At the risk of repetition, it should be stated again that in this case study, one feature is extracted from one sensor. 
A smaller number of neighbouring features means less redundant information in the selected features.
The average number of neighbouring features over the 16 tasks is 5.69 for the proposed algorithm, 6.38 for \texttt{SKB\_mic}, and 7.13 for \texttt{SKB\_clf}, respectively. The results illustrate that the proposed algorithm can better consider the information redundancy between features, thereby providing a feature set that is more effective for damage detection.}

\subsection{\RV{A case study on the edge computing application}}
\RV{Regarding the field deployment of SHM advanced algorithms, there are various approaches available, such as edge computing, cloud computing, and using industrial personal computers. 
Edge computing is recommended in this paper to implement the deployment of feature selection algorithms because it has the following advantages:
\begin{enumerate}
    \item It has local processing and storage capabilities that allow for uninterrupted operation and essential decision-making at the edge. 
    This is particularly valuable when deploying SHM systems where internet connectivity is unreliable, intermittent, or unavailable.
    \item It enables scalable and flexible architectures, where additional edge devices or nodes can be easily added or removed to meet the changing demands or requirements of an SHM project. 
    It also facilitates edge-cloud integration, supporting hybrid architectures that combine the benefits of both local processing and centralised cloud services.
    \item It can reduce the amount of data that needs to be transmitted to centralised cloud servers by processing data at the edge, leading to the reduced data transmission cost of an SHM system.
\end{enumerate}}

\begin{figure}[ht]
\centering
\includegraphics[width=0.45\linewidth]{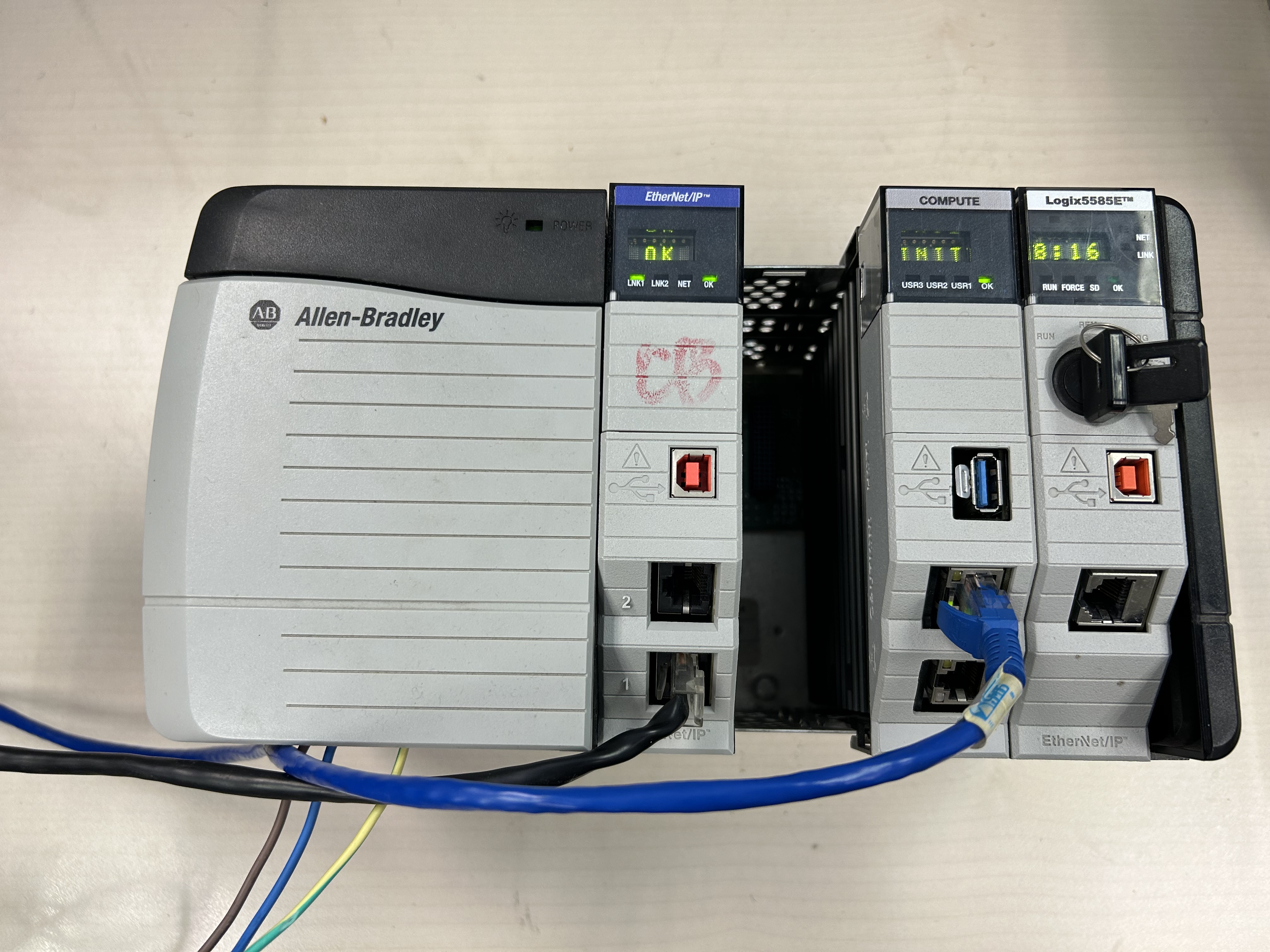}
\caption{Photograph of the edge computing module used.}
\label{fig:edge_compute}
\end{figure}

\RV{The Allen-Bradley\textsuperscript{\textregistered} ControlLogix\textsuperscript{\textregistered} 1756 Compute module, shown in the middle between EtherNet/IP\textsuperscript{TM} module and Logix5585E\textsuperscript{TM} module in Figure \ref{fig:edge_compute}, is used for demonstrating the computational speed performance of the proposed feature selection algorithm in an edge computing device.}

\begin{figure}[ht]
\centering
\begin{subfigure}[b]{\textwidth}
\centering
\includegraphics[width=1\linewidth]{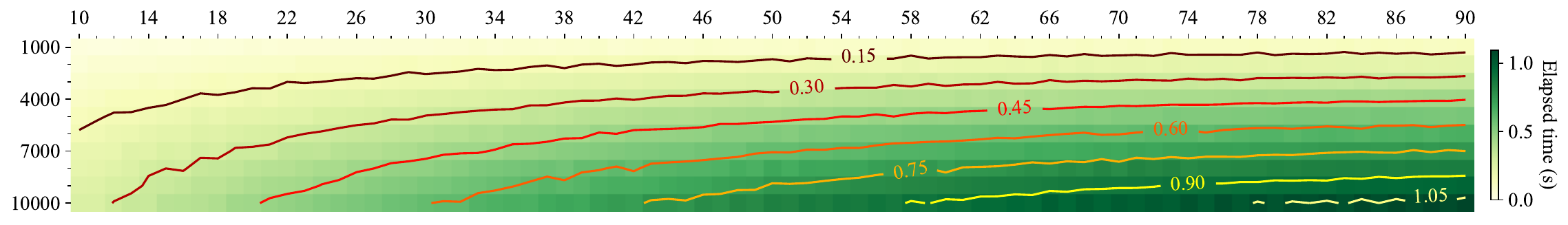}
\caption{\texttt{SSC\_h}}
\label{fig:edge_ssc_h}
\end{subfigure}
\hspace{1cm}
\begin{subfigure}[b]{\textwidth}
\centering
\includegraphics[width=1\linewidth]{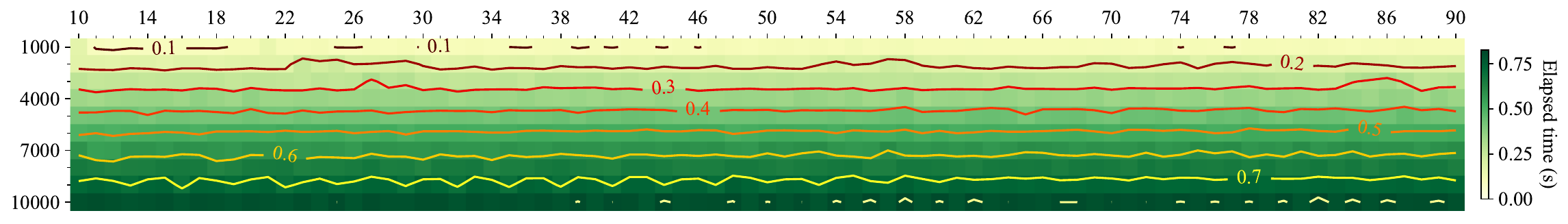}
\caption{\texttt{SSC\_eta}}
\label{fig:edge_ssc_eta}
\end{subfigure}

\begin{subfigure}[b]{0.32\textwidth}
\centering
\includegraphics[width=1\linewidth]{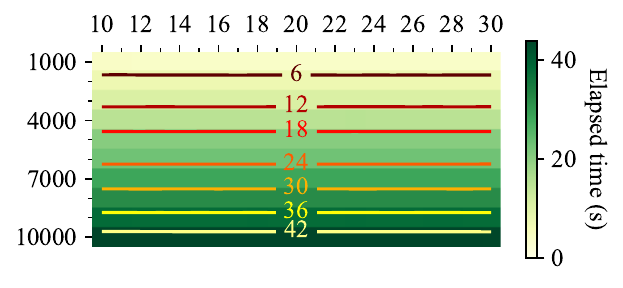}
\caption{\texttt{SKB\_mir}}
\label{fig:edge_skb_mir}
\end{subfigure}
\begin{subfigure}[b]{0.32\textwidth}
\centering
\includegraphics[width=1\linewidth]{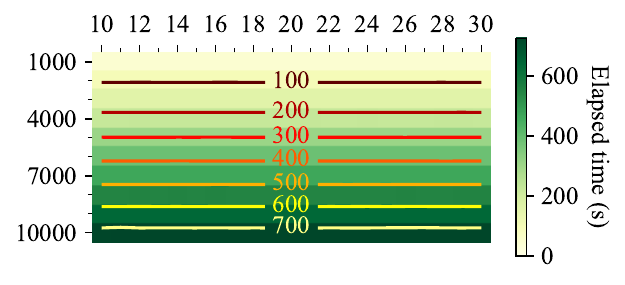}
\caption{\texttt{SFM\_rfr}}
\label{fig:edge_sfm_rfr}
\end{subfigure}
\begin{subfigure}[b]{0.32\textwidth}
\centering
\includegraphics[width=1\linewidth]{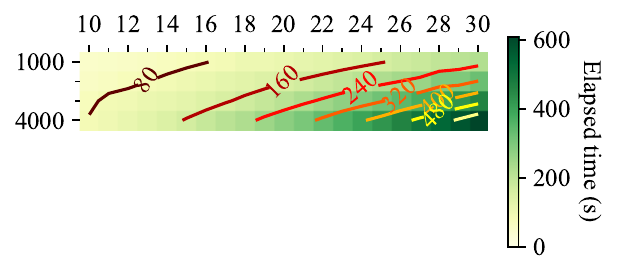}
\caption{\texttt{SFS\_lsvr}}
\label{fig:edge_sfs_lsvr}
\end{subfigure}

\caption{Computation time of different feature selection methods on the edge computing device. The $x$-axis represents the number of selected features and the $y$-axis represents the number of samples. The heat map (with colour bar) and contour lines represent the computation time. \texttt{SSC\_h} refers to the fast SSC-based feature selection algorithm accelerated with $h$-correlation, i.e. the $h$-correlation method. \texttt{SSC\_eta} refers to the fast SSC-based feature selection algorithm accelerated with $\eta$-cosine, i.e. the $\eta$-cosine method. \texttt{SKB\_mir} is a filter method. \texttt{SFM\_rfr} is an embedded method. \texttt{SFS\_lsvr} is a wrapper method.}
\label{fig:edge_time}
\end{figure}

\RV{In this case study, ten synthetic datasets are drawn from a standard normal distribution. Each of the datasets has 100 candidate features and one target variable. The sample sizes of the ten data sets range from 1000 to 10000, with a step size of 1000.
The computation time of the five feature selection methods is given in Figure \ref{fig:edge_time}, where
\begin{itemize}
    \item \texttt{SKB\_mir}, which is the SKB algorithm ranking features with mutual information for regression,
    \item \texttt{SFM\_rfr}, which is the SFM algorithm with a random forest regressor, and
    \item \texttt{SFS\_lsvr}, which is the forward SFS algorithm with a linear support vector regressor.
\end{itemize}
Since \texttt{SFS\_lsvr} is very slow to compute, feature selection is terminated when the sample size reaches 4000. However, the current results are sufficient for comparison and analysis of different methods.}

\RV{By comparing Figures \ref{fig:edge_ssc_h} and \ref{fig:edge_ssc_eta}, a similar phenomenon to that depicted in Figure \ref{fig:time} becomes evident. Specifically, when the number of selected features is increased up to a certain number, it is observed that the $\eta$-cosine method outperforms the $h$-correlation method in terms of computational efficiency.
Furthermore, in Figure \ref{fig:edge_time}, it can be observed that the elapsed time of both the $h$-correlation method and \texttt{SFS\_lsvr} increases as both the sample size and the number of selected features grow. In contrast, the $\eta$-cosine method, \texttt{SKB\_mir}, and \texttt{SFM\_rfr} show consistent computational efficiency regardless of the number of selected features.}

\RV{The second phenomenon can be attributed to the distinct selection strategies employed by the five methods. The filter method \texttt{SKB\_mir} and the embedded method \texttt{SFM\_rfr} select the top $t$ features based on a one-time computation of ranking criteria, i.e. mutual information for \texttt{SKB\_mir} and importance weights obtained from random forest models for \texttt{SFM\_rfr}. However, the $h$-correlation method and the wrapper method \texttt{SFS\_lsvr} need to compute the ranking criterion every time a feature is to be selected.}

\RV{Furthermore, although the $\eta$-cosine method also needs to compute the ranking criterion once per selection, it differs due to the benefit of a transformation step in Algorithm \ref{alg:ffs}, where the candidate features and target variable are transformed from the space $\mathbb{R}^{N \times 1 }$ to the space $\mathbb{R}^{101 \times 1}$ via the SVD, where $N$ represents the sample size. 
This transformation reduces the computation time required for subsequent feature selection in the space $\mathbb{R}^{101 \times 1}$, making the SVD computation time dominate the total computation time. Therefore, while the elapsed time increases along the $y$-axis (due to the increased computation time of SVD), its increase along the $x$-axis is not significant.}

\RV{In the context of edge computing, it can be inferred that the $h$-correlation method, the $\eta$-cosine method, and the filtering method \texttt{SKB\_mir} are more appropriate options because their computation speed is significantly faster than the embedded method \texttt{SFM\_rfr} and the wrapper method \texttt{SFS\_lsvr}. 
In addition, according to the analysis of Figure \ref{fig:SSCvsWrapper} and concrete examples in \ref{ap:d}, the proposed $h$-correlation and $\eta$-cosine methods are superior to the filtering method \texttt{SKB\_mir} in effectively addressing feature redundancy and interaction.
Therefore, it can be said that it is reasonable and effective to use the proposed SSC-based algorithm accelerated with $h$-correlation and $\eta$-cosine for feature selection tasks on the edge computing module.}

\section{Conclusions}
This paper proposes a fast feature selection algorithm based on the SSC, which can be used for SHM\RV{, as well as the general machine learning field.
Theorems have been developed to provide rigorous explanations for the computational speed enhancement of the $h$-correlation and $\eta$-cosine methods, as well as for their unification as the SSC-based feature selection algorithm. These theorems establish the formal basis for understanding the improved computational efficiency achieved by these two methods.}
The extraordinary speed advantage of the proposed methods for computing the SSC in greedy search has been demonstrated on the synthetic dataset.
The comparison between the eight feature ranking criteria shows that the SSC criterion can give competitive feature selection results for both general classification and regression tasks.

\RV{Moreover, three additional case studies demonstrate the application scenarios and performance of the proposed SSC-based feature selection algorithm in the SHM field.}
The comparison on the Gnat trainer dataset implies that the proposed feature selection algorithm has a similar performance to the embedded and wrapper methods in selecting damage-sensitive features but at a fast computational speed like filter methods.
\RV{The results of the glider wing dataset indicate that even in the presence of interference caused by environmental variations, the proposed algorithm can still maintain its advantages in selecting damage-sensitive features.}
\RV{The edge computing results} suggest that the proposed feature selection algorithm has great potential in SHM, especially where features need to be selected and updated online frequently, or where devices have limited computing capability.

\RV{Although the proposed feature selection algorithm performs well, two limitations need to be acknowledged.
Firstly, the proposed algorithm can only evaluate the linear relationship between features and output variables.
In addition, due to the greedy nature of the proposed algorithm, the selection results obtained are sub-optimal.
In future work, attempts will be made to alleviate these limitations using certain techniques, such as the kernel trick and beam search, to handle more complex real-world problems. 
More targeted usage scenarios for the proposed algorithm in the field of SHM will be further explored as well.}

\section*{Declaration of competing interest}
The authors declare that they have no known competing financial interests or personal relationships that could have appeared to influence the work reported in this paper.

\section*{Acknowledgements}
The authors would like to acknowledge the support for this work from the Engineering and Physical Sciences Research Council (EPSRC), UK, via grant numbers, EP/S001565/1 and EP/R004900/1, from Shanghai Qizhi Institute (SQI), China, via the grant number SQZ202310, \RV{and from Rockwell Automation Shanghai Research Center, China}.

\appendix
\section{Proof of the Correlation Superposition Theorem} \label{ap:a}
\begin{proof}
The canonical correlation coefficient on the left-hand side of equation \eqref{eq:c1} is defined in \eqref{eq:defcc}.
According to \eqref{eq:ccca}, the SSC, i.e. the sum of the eigenvalues, is given by,
\begin{equation} \label{eq:c1tr}
    \sum_{k = 1}^{n \wedge m}R_k^2({\mathbf{X},\mathbf{Y}}) = \tr{\left(\dotinv{\mathbf{X}_\mathrm{C}}{\mathbf{X}_\mathrm{C}}^{-1} \inp{\mathbf{X}_\mathrm{C}}{\mathbf{Y}_\mathrm{C}} \dotinv{\mathbf{Y}_\mathrm{C}}{\mathbf{Y}_\mathrm{C}}^{-1} \inp{\mathbf{Y}_\mathrm{C}}{\mathbf{X}_\mathrm{C}}\right)}
,\end{equation}
where the operator $\tr(\cdot)$ denotes a matrix trace.
According to the assumption \emph{\ref{s:c1.1}}, it can be found that $\mathbf{X}_\mathrm{C} = \mathbf{W}[\mathbf{X}_\mathrm{C}]_{\mathbf{W}}$, where $\mathbf{W} = \augm{\mathbf{W}_\mathrm{s}}{\mathbf{W}_\mathrm{r}}$.
Because of the similarity invariance of the trace, the right-hand side of \eqref{eq:c1tr} can then be rewritten as,
\begin{subequations} 
\begin{align}
    &\tr{\left(\dotinv{\mathbf{X}_\mathrm{C}}{\mathbf{X}_\mathrm{C}}^{-1} \inp{\mathbf{X}_\mathrm{C}}{\mathbf{Y}_\mathrm{C}} \dotinv{\mathbf{Y}_\mathrm{C}}{\mathbf{Y}_\mathrm{C}}^{-1} \inp{\mathbf{Y}_\mathrm{C}}{\mathbf{X}_\mathrm{C}}\right)}\\
    =&\tr{\left([\mathbf{X}_\mathrm{C}]_{\mathbf{W}} ^{-1}\dotinv{\mathbf{W}}{\mathbf{W}}^{-1} \inp{\mathbf{W}}{\mathbf{Y}_\mathrm{C}} \dotinv{\mathbf{Y}_\mathrm{C}}{\mathbf{Y}_\mathrm{C}}^{-1} \inp{\mathbf{Y}_\mathrm{C}}{\mathbf{W}} [\mathbf{X}_\mathrm{C}]_{\mathbf{W}} \right)}\\
    =&\tr{\left(\dotinv{\mathbf{W}}{\mathbf{W}}^{-1} \inp{\mathbf{W}}{\mathbf{Y}_\mathrm{C}} \dotinv{\mathbf{Y}_\mathrm{C}}{\mathbf{Y}_\mathrm{C}}^{-1} \inp{\mathbf{Y}_\mathrm{C}}{\mathbf{W}} \right)}\label{eq:c1tr1}\\
    =&\sum_{k = 1}^{n \wedge m}R_k^2({\mathbf{W},\mathbf{Y}})
.\end{align}
\end{subequations}
Due to the assumption \emph{\ref{s:c1.2}},
\begin{equation*}
\begin{split}
    \inp{\mathbf{W}}{\mathbf{W}} &= \begin{amatrix}{1}
    \inp{\mathbf{W}_\mathrm{s}}{\mathbf{W}_\mathrm{s}} & \mathbf{0}_{p\times q} \\
    \mathbf{0}_{q\times p} & \inp{\mathbf{W}_\mathrm{r}}{\mathbf{W}_\mathrm{r}}
    \end{amatrix} \\
    \inp{\mathbf{W}}{\mathbf{Y}_\mathrm{C}} &= \begin{pmatrix}
    \inp{\mathbf{W}_\mathrm{s}}{\mathbf{Y}_\mathrm{C}} \\
    \inp{\mathbf{W}_\mathrm{r}}{\mathbf{Y}_\mathrm{C}}
    \end{pmatrix} \\
    \inp{\mathbf{Y}_\mathrm{C}} {\mathbf{W}}&= \augm{\inp{\mathbf{Y}_\mathrm{C}}{\mathbf{W}_\mathrm{s}}}{\inp{\mathbf{Y}_\mathrm{C}}{\mathbf{W}_\mathrm{r}}}
.\end{split}
\end{equation*}
As a result of the columns of $\mathbf{W}_\mathrm{s}$, $\mathbf{W}_\mathrm{r}$ and $\mathbf{Y}_\mathrm{C}$ are zero-mean, it is found that \eqref{eq:c1tr1} is equal to
\begin{equation*}
\begin{split}
    &\tr{\left(\dotinv{\mathbf{W}_\mathrm{s}}{\mathbf{W}_\mathrm{s}}^{-1} \inp{\mathbf{W}_\mathrm{s}}{\mathbf{Y}_\mathrm{C}} \dotinv{\mathbf{Y}_\mathrm{C}}{\mathbf{Y}_\mathrm{C}}^{-1} \inp{\mathbf{Y}_\mathrm{C}}{\mathbf{W}_\mathrm{s}} \right)}+ \\
    &\tr{\left(\dotinv{\mathbf{W}_\mathrm{r}}{\mathbf{W}_\mathrm{r}}^{-1} \inp{\mathbf{W}_\mathrm{r}}{\mathbf{Y}_\mathrm{C}} \dotinv{\mathbf{Y}_\mathrm{C}}{\mathbf{Y}_\mathrm{C}}^{-1} \inp{\mathbf{Y}_\mathrm{C}}{\mathbf{W}_\mathrm{r}} \right)} \\
    =& \sum_{k = 1}^{p \wedge m}R_k^2({\mathbf{W}_\mathrm{s},\mathbf{Y}}) + \sum_{k = 1}^{q \wedge m}R_k^2({\mathbf{W}_\mathrm{r},\mathbf{Y}})
.\end{split}    
\end{equation*}
Therefore, Theorem \ref{c:1} follows.

\end{proof}

\section{Proof of Lemma \ref{lemma:1}} \label{ap:b}
\begin{proof}
According to the definition given by \eqref{eq:defcc}, the canonical correlation coefficient is given by,
\begin{subequations} 
\begin{align*}
    R_i({\mathbf{X},\mathbf{Y}}) &\coloneqq \max_{\bm{\upalpha}_i,\bm{\upbeta}_i} r({\mathbf{X}_\mathrm{C}\bm{\upalpha}_i,\mathbf{Y}_\mathrm{C}\bm{\upbeta}_i})\\
    & = \max_{\bm{\upalpha}_i,\bm{\upbeta}_i}\frac{\inp{(\mathbf{X}_\mathrm{C}\bm{\upalpha}_i)} {(\mathbf{Y}_\mathrm{C}\bm{\upbeta}_i)}}{\norm{\mathbf{X}_\mathrm{C}\bm{\upalpha}_i}\; \norm{\mathbf{Y}_\mathrm{C}\bm{\upbeta}_i}}
    \shortintertext{subject to}
    \inp{(\mathbf{X}_\mathrm{C}\bm{\upalpha}_i)}{(\mathbf{X}_\mathrm{C}\bm{\upalpha}_j)}&=0 \quad \text{for} \quad i\neq j,\\
    \inp{(\mathbf{Y}_\mathrm{C}\bm{\upbeta}_i)}{(\mathbf{Y}_\mathrm{C}\bm{\upbeta}_j)}&=0 \quad \text{for} \quad i\neq j  
,\end{align*} 
\end{subequations}
where $\bm{\upalpha}_i \in \mathbb{R}^{n\times 1}$ and $\bm{\upbeta}_i \in \mathbb{R}^{m\times 1}$.
From assumption \emph{\ref{s:l1.1}}, where,
\begin{subequations}
\begin{align*} 
    \mathbf{X}_\mathrm{C} &= \mathbf{U}[\mathbf{X}_\mathrm{C}]_{\mathbf{U}} \\
    \mathbf{Y}_\mathrm{C} &= \mathbf{U}[\mathbf{Y}_\mathrm{C}]_{\mathbf{U}} \\
    \inp{\mathbf{U}}{\mathbf{U}} &= \mathbf{I}_{n+m}
,\end{align*}
\end{subequations}
it can be seen that,
\begin{equation*}
\begin{split}
    \max_{\bm{\upalpha}_i,\bm{\upbeta}_i}\frac{\inp{(\mathbf{X}_\mathrm{C}\bm{\upalpha}_i)} {(\mathbf{Y}_\mathrm{C}\bm{\upbeta}_i})}{\norm{\mathbf{X}_\mathrm{C}\bm{\upalpha}_i}\; \norm{\mathbf{Y}_\mathrm{C}\bm{\upbeta}_i}}
    & = \max_{\bm{\upalpha}_i,\bm{\upbeta}_i}\frac{\inp{(\mathbf{U}[\mathbf{X}_\mathrm{C}]_{\mathbf{U}}\bm{\upalpha}_i)}{(\mathbf{U}[\mathbf{Y}_\mathrm{C}]_{\mathbf{U}}\bm{\upbeta}_i)}}{\norm{\mathbf{U}[\mathbf{X}_\mathrm{C}]_{\mathbf{U}}\bm{\upalpha}_i}\; \norm{\mathbf{U}[\mathbf{Y}_\mathrm{C}]_{\mathbf{U}}\bm{\upbeta}_i}} \\
    & = \max_{\bm{\upalpha}_i,\bm{\upbeta}_i}\frac{\inp{([\mathbf{X}_\mathrm{C}]_{\mathbf{U}}\bm{\upalpha}_i)} {([\mathbf{Y}_\mathrm{C}]_{\mathbf{U}}\bm{\upbeta}_i})}{\norm{[\mathbf{X}_\mathrm{C}]_{\mathbf{U}}\bm{\upalpha}_i}\; \norm{[\mathbf{Y}_\mathrm{C}]_{\mathbf{U}}\bm{\upbeta}_i}} \\
    & = \max_{\bm{\upalpha}_i,\bm{\upbeta}_i}\cos\left({\angle([\mathbf{X}_\mathrm{C}]_{\mathbf{U}}\bm{\upalpha}_i, [\mathbf{Y}_\mathrm{C}]_{\mathbf{U}}\bm{\upbeta}_i)}\right) \\
    & = \cos\left({\min_{\bm{\upalpha}_i,\bm{\upbeta}_i}\angle\left([\mathbf{X}_\mathrm{C}]_{\mathbf{U}}\bm{\upalpha}_i, [\mathbf{Y}_\mathrm{C}]_{\mathbf{U}}\bm{\upbeta}_i\right)}\right) \\
    & = \cos\big(\varTheta_i([\mathbf{X}_\mathrm{C}]_{\mathbf{U}}, [\mathbf{Y}_\mathrm{C}]_{\mathbf{U}})\big) \quad \text{for} \quad i=1,\ldots,n \wedge m
.\end{split}
\end{equation*}
Thus, Lemma \ref{lemma:1} is proved.

\end{proof}

\section{Proof of the Cosine Superposition Theorem} \label{ap:c}
\begin{proof}
The canonical correlation coefficient on the left-hand side of equation \eqref{eq:c2} is defined in \eqref{eq:defcc}.
According to \eqref{eq:ccca}, the SSC, i.e. the sum of the eigenvalues, is given by,
\begin{equation} \label{eq:c2tr_1}
    \sum_{k = 1}^{n \wedge m}R_k^2({\mathbf{X},\mathbf{Y}}) = \tr{\left(\dotinv{\mathbf{X}_\mathrm{C}}{\mathbf{X}_\mathrm{C}}^{-1} \inp{\mathbf{X}_\mathrm{C}}{\mathbf{Y}_\mathrm{C}} \dotinv{\mathbf{Y}_\mathrm{C}}{\mathbf{Y}_\mathrm{C}}^{-1} \inp{\mathbf{Y}_\mathrm{C}}{\mathbf{X}_\mathrm{C}}\right)}
,\end{equation}
According to the assumption \emph{\ref{s:c2.0}}, it can be found that 
$\mathbf{X}_\mathrm{C} = \mathbf{U}[\mathbf{X}_\mathrm{C}]_{\mathbf{U}}$ and $\mathbf{Y}_\mathrm{C} = \mathbf{U}[\mathbf{Y}_\mathrm{C}]_{\mathbf{U}}$, where 
$\inp{\mathbf{U}}{\mathbf{U}} = \mathbf{I}_{n+m}$. Thus, the right-hand side of (\ref{eq:c2tr_1}) can be rewritten as,
\begin{subequations} 
\begin{align}
    \sum_{k = 1}^{n \wedge m}R_k^2({\mathbf{X},\mathbf{Y}})
    &=\tr{\left(\dotinv{(\mathbf{U}[\mathbf{X}_\mathrm{C}]_{\mathbf{U}})}{\mathbf{U}[\mathbf{X}_\mathrm{C}]_{\mathbf{U}}}^{-1} \inp{(\mathbf{U}[\mathbf{X}_\mathrm{C}]_{\mathbf{U}})}{\mathbf{U}[\mathbf{Y}_\mathrm{C}]_{\mathbf{U}}}
    \dotinv{(\mathbf{U}[\mathbf{Y}_\mathrm{C}]_{\mathbf{U}})}{\mathbf{U}[\mathbf{Y}_\mathrm{C}]_{\mathbf{U}}}^{-1} \inp{(\mathbf{U}[\mathbf{Y}_\mathrm{C}]_{\mathbf{U}})}{\mathbf{U}[\mathbf{X}_\mathrm{C}]_{\mathbf{U}}}\right)}\\
    &=\tr{\left(\dotinv{[\mathbf{X}_\mathrm{C}]_{\mathbf{U}}}{[\mathbf{X}_\mathrm{C}]_{\mathbf{U}}}^{-1} \inp{[\mathbf{X}_\mathrm{C}]_{\mathbf{U}}}{[\mathbf{Y}_\mathrm{C}]_{\mathbf{U}}}
    \dotinv{[\mathbf{Y}_\mathrm{C}]_{\mathbf{U}}}{[\mathbf{Y}_\mathrm{C}]_{\mathbf{U}}}^{-1} \inp{[\mathbf{Y}_\mathrm{C}]_{\mathbf{U}}}{[\mathbf{X}_\mathrm{C}]_{\mathbf{U}}}\right)}\label{eq:c2tr_2}
.\end{align}
\end{subequations}
Note that according to Lemma \ref{lemma:1},
\begin{equation*}
    R_i(\mathbf{X}, \mathbf{Y}) = \cos\big(\varTheta_i([\mathbf{X}_\mathrm{C}]_{\mathbf{U}}, [\mathbf{Y}_\mathrm{C}]_{\mathbf{U}})\big)
.\end{equation*}
Therefore, \eqref{eq:c2tr_2} can also be obtained by substituting $[\mathbf{X}_\mathrm{C}]_{\mathbf{U}}$ for $\mathbf{X}$ and $[\mathbf{Y}_\mathrm{C}]_{\mathbf{U}}$ for $\mathbf{Y}$ in \eqref{eq:angle_eig}.

According to the assumption \emph{\ref{s:c2.1}}, it can be found that $[\mathbf{X}_\mathrm{C}]_{\mathbf{U}} = \mathbf{W}\big[[\mathbf{X}_\mathrm{C}]_{\mathbf{U}}\big]_{\mathbf{W}}$, where $\mathbf{W} = \augm{\mathbf{W}_\mathrm{s}}{\mathbf{W}_\mathrm{r}}$.
Because of the similarity invariance of the trace, the right-hand side of \eqref{eq:c2tr_2} can be rewritten as,
\begin{subequations} 
\begin{align}
    &\tr{\left(\dotinv{[\mathbf{X}_\mathrm{C}]_{\mathbf{U}}}{[\mathbf{X}_\mathrm{C}]_{\mathbf{U}}}^{-1} \inp{[\mathbf{X}_\mathrm{C}]_{\mathbf{U}}}{[\mathbf{Y}_\mathrm{C}]_{\mathbf{U}}} \dotinv{[\mathbf{Y}_\mathrm{C}]_{\mathbf{U}}}{[\mathbf{Y}_\mathrm{C}]_{\mathbf{U}}}^{-1} \inp{[\mathbf{Y}_\mathrm{C}]_{\mathbf{U}}}{[\mathbf{X}_\mathrm{C}]_{\mathbf{U}}}\right)}\\
    =&\tr{\left(\big[[\mathbf{X}_\mathrm{C}]_{\mathbf{U}}\big]_{\mathbf{W}} ^{-1}\dotinv{\mathbf{W}}{\mathbf{W}}^{-1} \inp{\mathbf{W}}{[\mathbf{Y}_\mathrm{C}]_{\mathbf{U}}} \dotinv{[\mathbf{Y}_\mathrm{C}]_{\mathbf{U}}}{[\mathbf{Y}_\mathrm{C}]_{\mathbf{U}}}^{-1} \inp{[\mathbf{Y}_\mathrm{C}]_{\mathbf{U}}}{\mathbf{W}} \big[[\mathbf{X}_\mathrm{C}]_{\mathbf{U}}\big]_{\mathbf{W}} \right)}\\
    =&\tr{\left(\dotinv{\mathbf{W}}{\mathbf{W}}^{-1} \inp{\mathbf{W}}{[\mathbf{Y}_\mathrm{C}]_{\mathbf{U}}} \dotinv{[\mathbf{Y}_\mathrm{C}]_{\mathbf{U}}}{[\mathbf{Y}_\mathrm{C}]_{\mathbf{U}}}^{-1} \inp{[\mathbf{Y}_\mathrm{C}]_{\mathbf{U}}}{\mathbf{W}} \right)}\label{eq:c2tr1}\\
    =&\sum_{k = 1}^{n \wedge m}\cos^2\big(\varTheta_k({\mathbf{W},[\mathbf{Y}_\mathrm{C}]_{\mathbf{U}}})\big)
.\end{align}
\end{subequations}
Because of the assumption \emph{\ref{s:c2.2}},
\begin{equation*}
\begin{split}
    \inp{\mathbf{W}}{\mathbf{W}} &= \begin{amatrix}{1}
    \inp{\mathbf{W}_\mathrm{s}}{\mathbf{W}_\mathrm{s}} & \mathbf{0}_{p\times q} \\
    \mathbf{0}_{q\times p} & \inp{\mathbf{W}_\mathrm{r}}{\mathbf{W}_\mathrm{r}}
    \end{amatrix} \\
    \inp{\mathbf{W}}{[\mathbf{Y}_\mathrm{C}]_{\mathbf{U}}} &= \begin{pmatrix}
    \inp{\mathbf{W}_\mathrm{s}}{[\mathbf{Y}_\mathrm{C}]_{\mathbf{U}}} \\
    \inp{\mathbf{W}_\mathrm{r}}{[\mathbf{Y}_\mathrm{C}]_{\mathbf{U}}}
    \end{pmatrix} \\
    \inp{[\mathbf{Y}_\mathrm{C}]_{\mathbf{U}}} {\mathbf{W}}&= \augm{\inp{[\mathbf{Y}_\mathrm{C}]_{\mathbf{U}}}{\mathbf{W}_\mathrm{s}}}{\inp{[\mathbf{Y}_\mathrm{C}]_{\mathbf{U}}}{\mathbf{W}_\mathrm{r}}}
.\end{split}
\end{equation*}
Then, it is found that \eqref{eq:c2tr1} is equal to
\begin{equation*}
\begin{split}
    &\tr{\left(\dotinv{\mathbf{W}_\mathrm{s}}{\mathbf{W}_\mathrm{s}}^{-1} \inp{\mathbf{W}_\mathrm{s}}{[\mathbf{Y}_\mathrm{C}]_{\mathbf{U}}} \dotinv{[\mathbf{Y}_\mathrm{C}]_{\mathbf{U}}}{[\mathbf{Y}_\mathrm{C}]_{\mathbf{U}}}^{-1} \inp{[\mathbf{Y}_\mathrm{C}]_{\mathbf{U}}}{\mathbf{W}_\mathrm{s}} \right)}+ \\
    &\tr{\left(\dotinv{\mathbf{W}_\mathrm{r}}{\mathbf{W}_\mathrm{r}}^{-1} \inp{\mathbf{W}_\mathrm{r}}{[\mathbf{Y}_\mathrm{C}]_{\mathbf{U}}} \dotinv{[\mathbf{Y}_\mathrm{C}]_{\mathbf{U}}}{[\mathbf{Y}_\mathrm{C}]_{\mathbf{U}}}^{-1} \inp{[\mathbf{Y}_\mathrm{C}]_{\mathbf{U}}}{\mathbf{W}_\mathrm{r}} \right)} \\
    =& \sum_{k = 1}^{p \wedge m}\cos^2\big(\varTheta_k(\mathbf{W}_\mathrm{s},[\mathbf{Y}_\mathrm{C}]_{\mathbf{U}})\big) + \sum_{k = 1}^{q \wedge m}\cos^2\big(\varTheta_k({\mathbf{W}_\mathrm{r},[\mathbf{Y}_\mathrm{C}]_{\mathbf{U}}})\big)
.\end{split}    
\end{equation*}
and Theorem \ref{c:2} follows.

\end{proof}

\section{\RV{Concrete examples of interaction and redundancy issues}}\label{ap:d}
\subsection{\RV{False interaction effect}}
\RV{Two features $\mathbf{x}_1 = (3, 0, 2, 1)^\top$ and $\mathbf{x}_2 = (0, 3, 1, 1)^\top$ are used here to construct a target variable $\mathbf{y}$, given by,
\begin{equation*}
    \mathbf{y} = \mathbf{x}_1+\mathbf{x}_2
\end{equation*}
Three candidate features, including $\mathbf{x}_1$, $\mathbf{x}_2$ and $\mathbf{x}_1 \odot \mathbf{x}_2$, are available to choose from to predict $\mathbf{y}$, where $\odot$ is the element-wise multiplication operator. When the absolute value of Pearson's correlation coefficient is used as the ranking criterion to evaluate a feature individually, the results are as follows,
\begin{equation*}
\begin{split}
    &\lvert r(\mathbf{x}_1, \mathbf{y})\rvert = 0.258 \\
    &\lvert r(\mathbf{x}_2, \mathbf{y})\rvert = 0.132 \\
    &\lvert r(\mathbf{x}_1 \odot \mathbf{x}_2, \mathbf{y})\rvert = 0.174
.\end{split}    
\end{equation*}
Then, features $\mathbf{x}_1$ and $\mathbf{x}_1 \odot \mathbf{x}_2$ will be selected to predict $\mathbf{y}$, which means the interaction term  $\mathbf{x}_1 \odot \mathbf{x}_2$ is falsely selected.}

\RV{The proposed SSC-based feature selection algorithm evaluates a feature subset rather than an individual feature. Therefore, when the proposed algorithm is adopted, after selecting the first feature $\mathbf{x}_1$,  $R^2((\mathbf{x}_1 | \mathbf{x}_2), \mathbf{y}) = 1$ and $R^2((\mathbf{x}_1 | \mathbf{x}_1 \odot \mathbf{x}_2), \mathbf{y}) = 0.111$ are then obtained, and $\mathbf{x}_2$ will be selected as the second feature. The result indicates that the proposed algorithm is immune to the false interaction.}

\subsection{\RV{True interaction effect}}
\RV{Given two features $\mathbf{x}_1 = (1, 0, 0, 1)^\top$, $\mathbf{x}_2 = (0, 2, 3, -1)^\top$ and their interaction term $\mathbf{x}_1 \odot \mathbf{x}_2 = (0, 0, 0, -1)^\top$, a target variable $\mathbf{y}$ is constructed by,
\begin{equation*}
    \mathbf{y} = \mathbf{x}_1+\mathbf{x}_1 \odot \mathbf{x}_2
\end{equation*}
When the absolute value of Pearson's correlation coefficient is used as the ranking criterion to evaluate three candidate features ($\mathbf{x}_1$, $\mathbf{x}_2$ and $\mathbf{x}_1 \odot \mathbf{x}_2$) independently, the results are as follows,
\begin{equation*}
\begin{split}
    &\lvert r(\mathbf{x}_1, \mathbf{y})\rvert = 0.577 \\
    &\lvert r(\mathbf{x}_2, \mathbf{y})\rvert = 0.365 \\
    &\lvert r(\mathbf{x}_1 \odot \mathbf{x}_2, \mathbf{y})\rvert = 0.333
.\end{split}    
\end{equation*}
Then, features $\mathbf{x}_1$ and $\mathbf{x}_2$ will be selected to predict $\mathbf{y}$, which means the interaction term  $\mathbf{x}_1 \odot \mathbf{x}_2$ is falsely missed. When the proposed SSC-based algorithm is adopted, after selecting the first feature $\mathbf{x}_1$, $R^2((\mathbf{x}_1 | \mathbf{x}_2), \mathbf{y}) = 0.667$ and $R^2((\mathbf{x}_1 | \mathbf{x}_1 \odot \mathbf{x}_2), \mathbf{y}) = 1$ are then obtained, and $\mathbf{x}_1 \odot \mathbf{x}_2$ will be selected as the second feature. The result indicates that the proposed algorithm can successfully select the true interaction term.}

\subsection{\RV{Colinear redundancy}}
\RV{Given three features $\mathbf{x}_1 = (1, 1, 0, 0)^\top$, $\mathbf{x}_2 = (1, 2, 0, 0)^\top$ and $\mathbf{x}_3 = (1, 0, 0, 1)^\top$, a target variable $\mathbf{y}$ is constructed by,
\begin{equation*}
   \mathbf{y} = \mathbf{x}_1+\mathbf{x}_2+\mathbf{x}_3 
\end{equation*}
When the absolute value of Pearson's correlation coefficient is used as the ranking criterion to evaluate four candidate features (i.e., $\mathbf{x}_1$, $\mathbf{x}_2$, $\mathbf{x}_3$ and a colinear feature $\mathbf{x}_1-\mathbf{x}_2 = (0, -1, 0, 0)^\top$) independently, the results are given by,   
\begin{equation*}
\begin{split}
    &\lvert r(\mathbf{x}_1, \mathbf{y})\rvert = 0.962 \\
    &\lvert r(\mathbf{x}_2, \mathbf{y})\rvert = 0.870 \\
    &\lvert r(\mathbf{x}_3, \mathbf{y})\rvert = 0.192 \\
    &\lvert r(\mathbf{x}_1 - \mathbf{x}_2, \mathbf{y}) \rvert= 0.556
.\end{split}    
\end{equation*}
It can be seen that the redundant feature $\mathbf{x}_1 - \mathbf{x}_2$ is preferred over $\mathbf{x}_3$, which means features $\mathbf{x}_1$, $\mathbf{x}_2$ and $\mathbf{x}_1 - \mathbf{x}_2$ will be selected to predict $\mathbf{y}$. The colinear feature $\mathbf{x}_1 - \mathbf{x}_2$ is incorrectly selected here.}

\RV{It is worth mentioning that there are two ways for the proposed algorithm to deal with redundant features. The first way is implemented by evaluating a subset of features, similar to how the algorithm handles interactions between features. 
Suppose a candidate feature contains redundant information to the selected features. In that case, its contribution to the SSC will be small when combined with these features, and then it will be excluded by the proposed algorithm.
The second way is achieved by utilising the rounding error of the modified Gram-Schmidt orthogonalisation to skip the colinear features, as shown in Line \ref{alg:ldc} of Algorithm \ref{alg:ffs}. The rounding error is expected to be zero after the modified Gram-Schmidt orthogonalisation process.}

\RV{The second way is applied to this concrete example to recognise the colinear feature. After selecting features $\mathbf{x}_1$ and $\mathbf{x}_2$, the proposed algorithm will make the remaining candidate features orthogonal to $\mathbf{x}_1$ and $\mathbf{x}_2$. 
The rounding error for $\mathbf{x}_3$ is $3.722\times e^{-16}$, which is close to zero. However, the absolute value of the rounding error for $\mathbf{x}_1 - \mathbf{x}_2$ is $0.924$, which is significantly greater than zero. 
Therefore, the colinear feature $\mathbf{x}_1 - \mathbf{x}_2$ will be easily recognised and excluded.}

\bibliographystyle{unsrt}
\bibliography{ref}

\end{document}